\documentclass[10pt,twocolumn,letterpaper]{article}

\usepackage{wacv}
\usepackage{times}
\usepackage{epsfig}
\usepackage{graphicx}
\usepackage{amsmath}
\usepackage{amssymb}
\usepackage{subfigure}
\usepackage{booktabs}
\usepackage{multirow}
\usepackage{url}
\usepackage[T1]{fontenc}
\usepackage[utf8]{inputenc}
\usepackage{authblk}

\wacvfinalcopy 

\def\wacvPaperID{22} 

\ifwacvfinal\fi
\begin{document}

\title{ReMotENet: Efficient Relevant Motion Event Detection \\ for
Large-scale Home Surveillance Videos} 


\author{Ruichi Yu$^{1,2}$ ~~~~~Hongcheng Wang$^2$~~~~~Larry S. Davis$^1$\\
$^1$University of Maryland, College Park ~~~~~~~~~~$^2$Comcast Applied AI Research, D.C. \\
{\tt\small {\{richyu, lsd\}@umiacs.umd.edu~~~Hongcheng\_Wang@comcast.com}
}
}


\maketitle
\ifwacvfinal\thispagestyle{empty}\fi

\begin{abstract}
This paper addresses the problem of detecting relevant motion caused by objects of interest (e.g., person and vehicles) in large scale home surveillance videos. The traditional method usually consists of two separate steps, i.e., detecting moving objects with background subtraction running on the camera, and filtering out nuisance motion events with deep learning based object detection and tracking running on cloud. The method is extremely slow, and does not fully leverage the spatial-temporal redundancies with a pre-trained off-the-shelf object detector. To dramatically speedup relevant motion event detection and improve its performance, we propose a novel network for relevant motion event detection, ReMotENet, which is a unified, end-to-end data-driven method using spatial-temporal attention-based 3D ConvNets to jointly model the appearance and motion of objects-of-interest in a video. ReMotENet parses an entire video clip in one forward pass of a neural network to achieve significant speedup, which exploits the properties of home surveillance videos, and enhances 3D ConvNets with a spatial-temporal attention model and frame differencing to encourage the network to focus on the relevant moving objects. Experiments demonstrate that our method can achieve comparable or event better performance than the object detection based method but with three to four orders of magnitude speedup (up to 20k$\times$) on GPU devices. Our network is efficient, compact and light-weight. It can detect relevant motion on a 15s surveillance video clip within 4-8 milliseconds on a GPU and a fraction of second (0.17-0.39s) on a CPU with a model size of less than 1MB.
\end{abstract}

\section{Introduction}
With the development of home security and surveillance system, more and more home surveillance cameras have been installed to monitor customers' home 24/7 for security and safety purpose. Most existing commercial solutions run motion detection on the edge (camera), and show the detected motion events (usually in short clips of, e.g., 15s) for end users' review on the web or mobile.

Motion detection has been a challenging problem in spite of many years' development in academia and industry \cite{motiondet2014,bgslibrarychapter}. Nuisance alarm sources, such as tree motion, shadows, reflections, rain/snow, flags, result in many irrelevant motion events for customers. 

Relevant motion detection is responsive to customers' needs. It involves pre-specified relevant objects, e.g., people, vehicles and pets, and these objects should have human recognizable location changes in the video. 
It not only helps to remove nuisance events, but also supports applications such as semantic video search and video summarization. 

\begin{figure}[!t]
\centering
  \includegraphics[height=4cm,width=8cm]{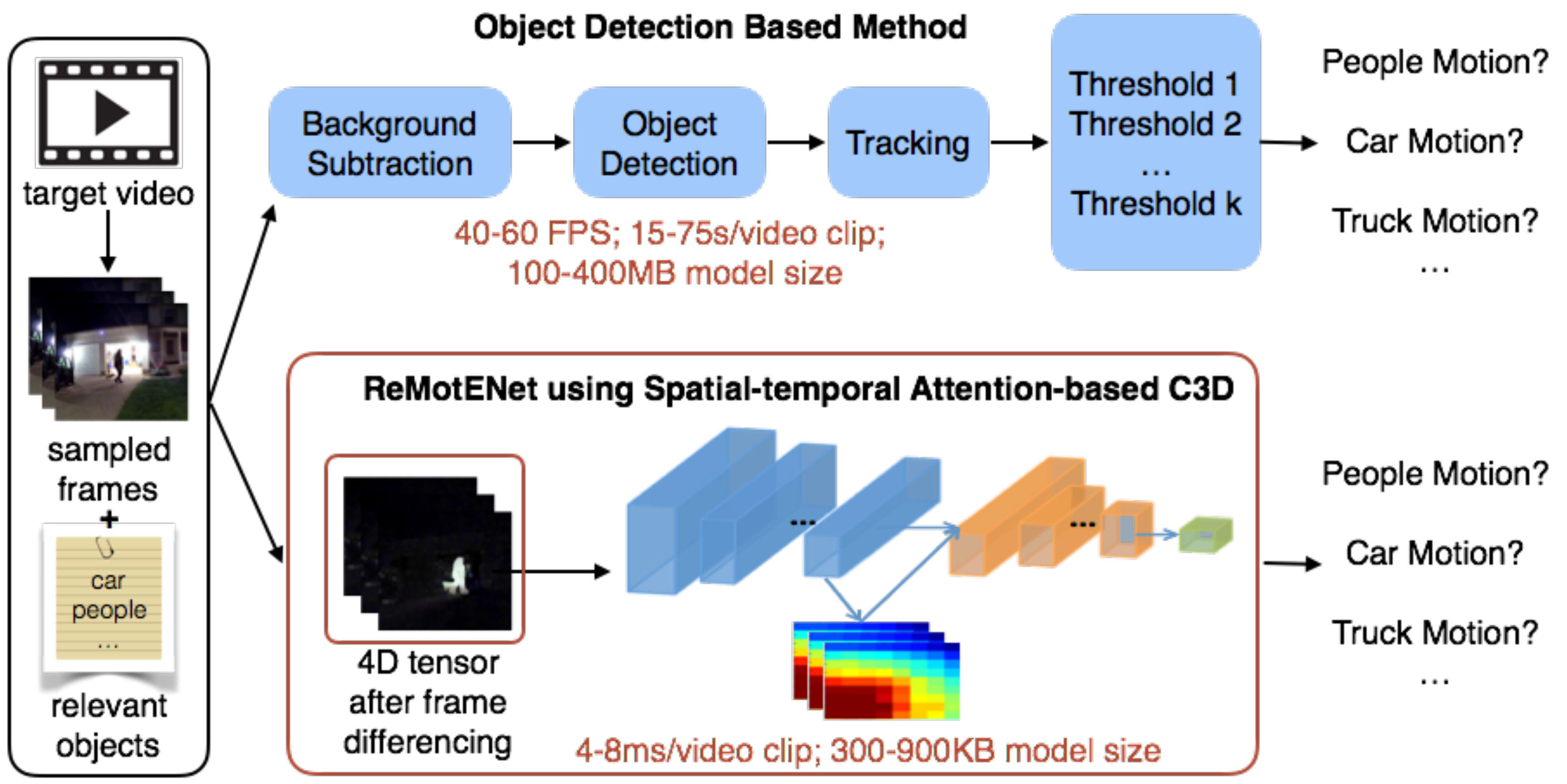}
  \caption{Comparison between the traditional object detection based method and ReMotENet: Traditional method conducts background subtraction, object detection and tracking on each frame; ReMotENet parses a 4D representation of the entire video clip in one forward pass of 3D ConvNets to efficiently detect relevant motion involving objects-of-interest.}
\label{fig:intro}
\end{figure}

As shown in Figure \ref{fig:intro}, one natural method is to apply state-of-the-art object detectors based on deep convolutional neural networks (CNNs) \cite{RCNN,fastRcnn,faster,SSD,YOLO,yolo2} to identify objects of interest. 
Given a video clip, background subtraction is applied to each frame to filter out stationary frames. Object detection is then applied to frames that have motion to identify the moving objects. Finally, the system generates track from the detection results to filter out temporally inconsistent false detections or stationary ones.

There are at least two problems with this object detection based method. First, it is computationally expensive, especially the object detection module. The state-of-the-art object detectors \cite{RCNN,fastRcnn,faster,SSD,YOLO,yolo2} need to be run on expensive GPUs devices and achieve at most 40-60 FPS \cite{API}. Scaling to tens of thousands of motion events coming from millions of cameras becomes cost ineffective. Second, the method usually consists of several separate pre-trained methods or hand-crafted rules, and does not fully utilize the spatial-temporal information of an entire video clip. For example, moving object categories are detected mainly by object detection, which ignores motion patterns that can also be utilized to classify the categories of moving objects.

To address these problems, we propose a network for relevant motion event detection, ReMotENet, which is a unified, end-to-end data-driven method using Spatial-temporal Attention-based 3D ConvNets to jointly model the appearance and motion of objects-of-interest in a video event. As shown in Figure \ref{fig:intro}, ReMotENet parses an entire video clip in one forward pass of a neural network to achieve significant speedup (up to 20k$\times$) on a single GPU. This makes the system easily scalable to millions of motion events and reduces latency. Meanwhile, it exploits the properties of home surveillance videos, e.g., relevant motion is sparse both spatially and temporally, and enhances 3D ConvNets with a spatial-temporal attention model and frame differencing to encourage the network to focus on relevant moving objects. 

To train and evaluate our model, we collected a large dataset of 38,360 real home surveillance video clips of 15s from 78 cameras covering various scenes, including indoor/outdoor, day/night, different lighting conditions and weather. To avoid the cost of training annotations, our method is weakly supervised by the results of the object detection based method. For evaluation, we manually annotated 9,628 video clips with binary labels of relevant motion caused by different objects. 
Experiments demonstrate that ReMotENet achieves comparable or even better performance, but is three to four orders of magnitude faster than the object detection based method. Our network is efficient, compact and light-weight. It can precisely detect relevant motion in a 15s video in 4-8 milliseconds on a GPU and a fraction of second on a CPU with model size of less than 1MB. 


\section{Related Work}
Our application is to detect relevant motion events in surveillance videos. As far as we know, there is no existing work addressing the relevant motion detection application by unifying motion detection and objects of interest detection. Most relevant work usually addresses this problem with a two-step detection based approach consisting of detecting moving objects with background subtraction and filtering out nuisance motion events with object detection and tracking.


Background subtraction has been widely used and studied \cite{motiondet2014,bgslibrarychapter,BGS-G,BGS-GMM1,background} to detect moving objects from videos. 
It extracts motion blobs but does not recognize the semantic categories of the moving objects, and many nuisance alarm sources
result in irrelevant motion events. 

Semantic object detection is then applied to further filter out these nuisance alarms. One of the state-of-the-art object detection frameworks is R-CNN and its variants \cite{RCNN,fastRcnn,faster}. Recently, more efficient detectors, e.g., YOLO \cite{YOLO,yolo2} and SSD \cite{SSD} have been proposed to speedup the detection pipeline with some performance degradation compared to R-CNN.
To identify moving objects in a video, tracking techniques (traditional \cite{sort,track1,track2} and deep network based \cite{track-nn1,track-nn2,track-nn3}) are usually used to further reduce false detections. 
The detection based approach is computationally expensive. Significant GPU resources are needed when we consider millions of surveillance cameras. Meanwhile,
some hand-crafted and ad-hoc hyper-parameters or thresholds (e.g., minimum motion object size, the detection confidence threshold and length of valid tracker threshold) are needed to model relevant motion. In addition, one cannot fully leverage the spatial-temporal redundancies with just a pre-trained off-the-shelf object detector and a frame-by-frame detection.

Our work is also related to video activity recognition, which is to detect and categorize activities (usually human activities) in videos. To model motion and temporal information in a video, two-stream networks \cite{2stream}, long-term recurrent neural networks \cite{lstm} and 3D convolution networks (3D ConvNets) based methods \cite{C3D} have been proposed. 
Our task is different from the video activity recognition task in two aspects: first, we only care about the categories of moving objects, rather than fine-grained categories of the activities of one specific object class; second, due to the large volume of videos, small computational cost has higher priority in our application. 

Recently, Kang \emph{et al.} \cite{noscope} proposed NoScope, a pre-processing method to reduce the number of frames needed to be parsed in an object-detection based video query system. They utilize frame difference and specialized networks to filter out frames without moving relevant objects. There are several major difference between NoScope and ReMotENet: 
{First}, ReMotENet is a unified end-to-end solution without explicit object detection, but NoScope is a pre-processing step for object detection which combines several pre-trained models. 
Although our method is weakly supervised by object detection, it learns the general motion and appearance patterns of different objects from the noisy labels and recovers from some mistakes made by the detection pipeline. 
However, as a pre-processing step, NoScope highly relies on the performance of pre-trained object detectors, which can be unreliable, especially on home surveillance videos with low video quality, lighting changes and various weather conditions.
{Second}, ReMotENet jointly models frames in a video clip, but NoScope conducts detection independently in a frame-by-frame fashion. Since NoScope relies on object detectors, the run-time speedup is insignificant. From the results shown in \cite{noscope}, when the run-time speedup is $>$ 100$\times$, the performance of NoScope drops quickly. However, our method can achieve more than 19k$\times$ speedup while achieving comparable or better performance than the object detection baseline.
We do not quantitatively compare to NoScope mainly because the two methods are designed for different tasks: NoScope focuses on frame-level object detection while ReMotENet is designed for video-clip-level relevant motion detection.

\section{Our Approach}

\begin{figure}[!t]
\centering
  \includegraphics[height=3.5cm,width=8cm]{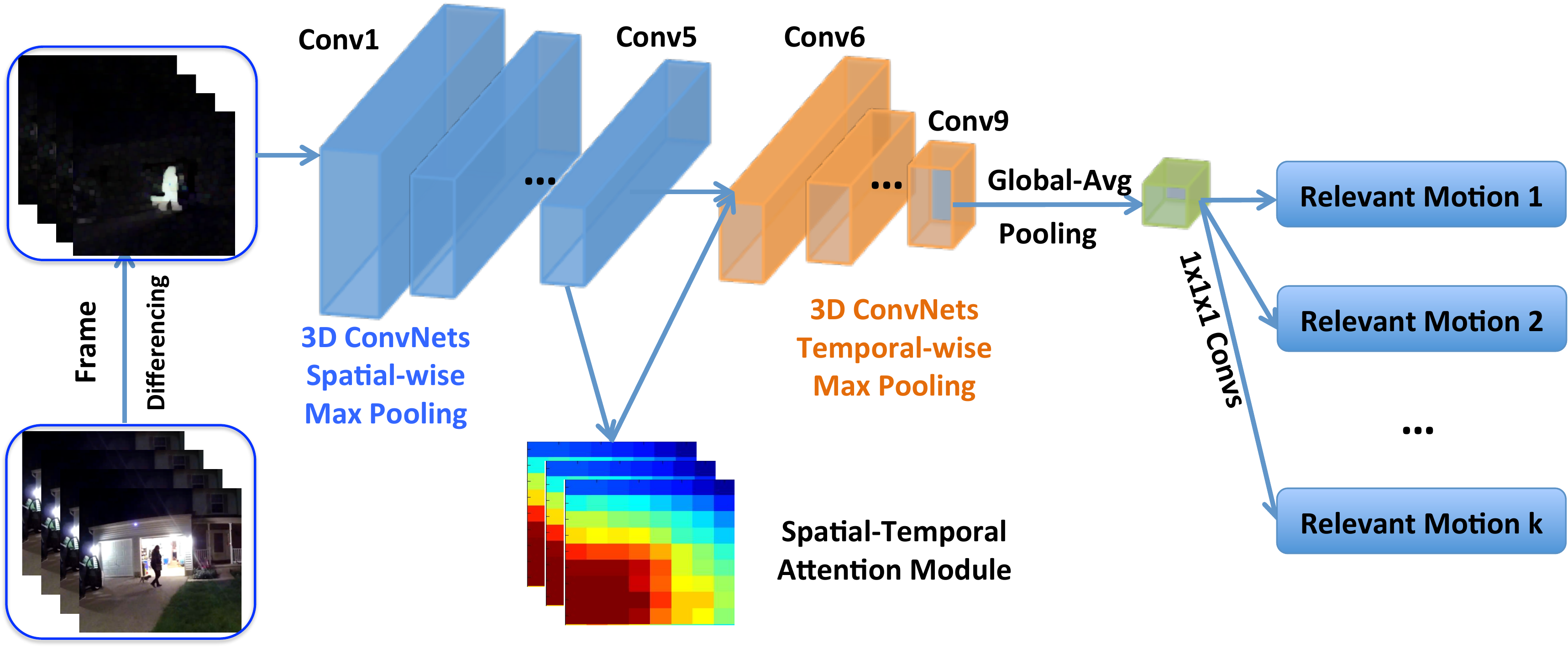}
  \caption{ReMotENet for Relevant Motion Event Detection. The input is a 4D representation of an entire video clip and the outputs are several binary predictions of relevant motion involving different objects-of-interest. The low-level 3D ConvNets in blue only abstract spatial features with spatial-wise max pooling. The high-level 3D ConvNets in orange abstract temporal features using temporal-wise max pooling. We predict a spatial-temporal mask and multiply it with the extracted features from Conv5 (with Pool5) before it is fed as the input to Conv6.}
\label{fig:sys}
\end{figure}



To dramatically speedup relevant motion event detection and improve its performance, instead of directly compressing deep network \cite{NISP,prune}, we propose a novel network for relevant motion event detection, ReMotENet, which is a unified, end-to-end data-driven method using spatial-temporal attention-based 3D ConvNets to jointly model the appearance and motion of objects-of-interest in a video. 

\subsection{Analyzing an Entire Video Clip at Once}
In contrast to an object detection based method based on frame-by-frame processing, we propose a unified, end-to-end data-driven framework that takes an entire video clip as input to detect relevant motion using 3D ConvNets \cite{C3D}.

One crucial advantage of using 3D ConvNets 
is that they can parse an entire video clip in one forward pass of a deep network, which is extremely efficient. Meanwhile, unlike the traditional pipeline that conducts object detection and tracking independently, a 3D ConvNet is an end-to-end model that jointly models the appearance of objects and their motion patterns.
To fit an entire video in memory, we down-sample the video frames spatially and temporally.
We sample one FPS to uniformly sample 15 frames from a 15s video clip, and reduce the original resolution (1280$\times$720) by a factor of 8. 
The input tensor of our 3D ConvNets is 15$\times$90$\times$160$\times$3.

\subsection{ReMotENet using Spatial-temporal Attention-based C3D}

\subsubsection{Frame Differencing}
Global or local context of both background or foreground objects has proven to be useful for activity recognition\cite{context_act1,context_act3}
(e.g., some sports only happen on playgrounds; some collective activities have certain spatial arrangements of the objects that participate). 
However, since home surveillance cameras capture different scenes at different times with various weather and lighting conditions, the same relevant motion could involve different background and foreground arrangements. 

So, we conduct a simple and efficient frame differencing on the 4D input tensor to suppress the influence of the background and foreground variance: 
for each frame sub-sampled from a video clip, we select its previous frame as a ``reference-frame" and difference them. 
The image quality of surveillance cameras is usually low, and the moving objects are relatively small due to large field of view. As a result, it requires high resolution videos for object detector to capture fine-grained features, such as texture, to detect relevant objects.
However, for ReMotENet with frame differencing, since we skip explicit object detection, it is sufficient for the network to learn coarse appearance features of objects, e.g., shape, contour and aspect-ratio with frames of low resolution, which leads to significant speedup. 


\subsubsection{Spatial-temporal Attention-based 3D ConvNets}
Most video clips captured by a home surveillance camera may only contain the stationary scene with irrelevant motion such as shadow, rain and parked vehicles. Meanwhile, our network should be capable of differentiating appearance and motion pattern of different objects-of-interest.

Besides using frame differencing to suppress the background, we propose a Spatial-Temporal Attention-based (STA) model as shown in Figure \ref{fig:sys} to guide the network to only pay attention to the moving objects of interest and filter out foreground motion caused by irrelevant objects. 
Unlike most of the attention models that are trained end-to-end with the groundtruth of final predictions, 
we use the detection confidence scores and bounding boxes of the moving relevant objects obtained from a general detector (e.g., Faster R-CNN) as pseudo-groundtruth to train the STA model. 

The supervision from the detection results suppresses the influence of features from irrelevant regions of each frame, and the binary labels of motion adjusts the STA layer to pay attention to the spatial-temporal locations that are most informative for predicting the relevant motion. The spatial-temporal attention model can be viewed as a combination of multi-task learning and a traditional attention model. In the weakly supervised framework, it helps our network to learn from noisy labels and recover from some of the mistakes caused by the object detection based method. For instance, some relevant motion is missed because of bad thresholds of the tracking module, but the object detector has correctly detected the relevant objects. The STA model trained with those detection results allows our network to recover from the binary prediction mistake. 

In contrast to the original C3D model that conducts max pooling both spatially and temporally \cite{C3D}, we separate the spatial and temporal max pooling as shown in Figure \ref{fig:sys}. This allows us to generate an attention mask on each input frame to capture fine-grained temporal information, and makes the network deeper to learn better representations. 
We first apply five layers of 3D convolutions (Conv1-Conv5) with spatial max pooling on the 4D input tensor after frame differencing to extract the appearance based features.
Then, we apply another 3D convolution layer (STA layer) on the output of Pool5 to obtain a binary prediction of whether whether to attend each spatial-temporal location. 
We conduct a softmax operation on the binary predictions to compute a soft probability of attention for each spatial-temporal location. We scale the features from Pool5 by applying an element-wise multiplication between the attention mask and the extracted features. 
Another four layers of 3D ConvNets are used with temporal max pooling to abstract temporal features. 

The network in \cite{C3D} utilized several fully connected layers after the last convolution layer, which leads to a huge number of parameters and computations.
Inspired by \cite{google}, we apply a spatial global average pooling (GAP) to aggregate spatial features after Pool9 and use several 1$\times$1$\times$1 convolution layers with two filters (denoted as ``Binary" layers) to predict the final binary results. The use of GAP and 1$\times$1$\times$1 convolutions significantly reduces the number of parameters and model size of our method. 
The final outputs of our network are several binary predictions indicating whether there is any relevant motion of a certain object or a group of objects. The detailed network structure is shown in Table \ref{network}. 
For each Conv layer, we use ReLU as its activation.

\begin{table}[!h]
\centering
\scriptsize
\caption{Network Structure of the ReMotENet using Spatial-temporal Attention-based 3D ConvNets}
\label{network}
\begin{tabular}{@{}c|c|c|c|c@{}}
\toprule
Layer     & Input Size  & Kernel Size & Stride & Num of Filters \\ \midrule
Conv1     & 15$\times$90$\times$160$\times$3 & 3$\times$3$\times$3       & 1$\times$1$\times$1  & 16      \\
Pool1     & 15$\times$90$\times$160$\times$3 & 1$\times$2$\times$2       & 1$\times$2$\times$2  & -       \\
Conv2     & 15$\times$45$\times$80$\times$16 & 3$\times$3$\times$3       & 1$\times$1$\times$1  & 16      \\
Pool2     & 15$\times$45$\times$80$\times$16 & 1$\times$2$\times$2       & 1$\times$2$\times$2  & -       \\
Conv3     & 15$\times$23$\times$40$\times$16 & 3$\times$3$\times$3       & 1$\times$1$\times$1  & 16      \\
Pool3     & 15$\times$23$\times$40$\times$16 & 1$\times$2$\times$2       & 1$\times$2$\times$2  & -       \\
Conv4     & 15$\times$12$\times$20$\times$16 & 3$\times$3$\times$3       & 1$\times$1$\times$1  & 16      \\
Pool4     & 15$\times$12$\times$20$\times$16 & 1$\times$2$\times$2       & 1$\times$2$\times$2  & -       \\
Conv5     & 15$\times$6$\times$10$\times$16  & 3$\times$3$\times$3       & 1$\times$1$\times$1  & 16      \\
Pool5     & 15$\times$6$\times$10$\times$16  & 1$\times$2$\times$2       & 1$\times$2$\times$2  & -       \\ \midrule
STA      & 15$\times$3$\times$5$\times$16   & 3$\times$3$\times$3       & 1$\times$1$\times$1  & 2       \\ \midrule
Conv6     & 15$\times$3$\times$5$\times$16   & 3$\times$3$\times$3       & 1$\times$1$\times$1  & 16      \\
Pool6     & 15$\times$3$\times$5$\times$16   & 2$\times$1$\times$1       & 2$\times$1$\times$1  & -       \\
Conv7     & 8$\times$3$\times$5$\times$16    & 3$\times$3$\times$3       & 1$\times$1$\times$1  & 16      \\
Pool7     & 8$\times$3$\times$5$\times$16    & 2$\times$1$\times$1       & 2$\times$1$\times$1  & -       \\
Conv8     & 4$\times$3$\times$5$\times$16    & 3$\times$3$\times$3       & 1$\times$1$\times$1  & 16      \\
Pool8     & 4$\times$3$\times$5$\times$16    & 2$\times$1$\times$1       & 2$\times$1$\times$1  & -       \\
Conv9     & 2$\times$3$\times$5$\times$16    & 3$\times$3$\times$3       & 1$\times$1$\times$1  & 16      \\
Pool9     & 2$\times$3$\times$5$\times$16    & 2$\times$1$\times$1       & 2$\times$1$\times$1  & -       \\
GAP       & 1$\times$3$\times$5$\times$16    & 1$\times$3$\times$5       & 1$\times$1$\times$1  & -       \\
Binary & 1$\times$1$\times$1$\times$16    & 1$\times$1$\times$1           &  1$\times$1$\times$1       & 2      \\ \bottomrule
\end{tabular}
\end{table}

\subsection{Network Training}
Considering the large volume of home surveillance videos, it is time-consuming to annotate each training video with binary labels. So, we adopt a weakly-supervised learning framework that utilizes the pseudo-groundtruth generated from the object detection based method (details are discussed in section \ref{dataset}).
%
Besides binary labels, we also utilize the pseudo-groundtruth of detection confidence scores and bounding boxes of moving objects-of-interest obtained from the object detection based method to train the STA layer.
The loss function of the STA layer is:
\begin{align}
\label{loss}
  loss =  &  \frac{1}{N} \biggr( C_1\sum_n \sum_i w_{n,i} \text{CE} (g_{n,i},y_{n,i}) \\ \nonumber
    &  + \frac{C_2}{ W\cdot H \cdot T} \sum_n \sum_{w,h,t}\text{CE}(sta_{n,w,h,t},Gsta_{n,w,h,t}) \biggr)
\end{align}

The first part is the cross-entropy loss (CE) for each relevant motion category; the second part is the CE loss between the predicted attention of each spatial-temporal location produced by ``STA" layer and the pseudo-groundtruth obtained from the object detector. $W,H,T$ are spatial and temporal sizes of the responses of layer ``STA"; $y_{n,i}$ and $g_{n,i}$ are the predicted and groundtruth motion labels of the $n^{th}$ sample; $sta_{n,w,h,t}$ and $Gsta_{n,w,h,t}$ are the predicted and groundtruth attention probabilities; $w_{n,i}$ is the loss weight of the $n^{th}$ sample, which is used to balance the biased number of positive and negative training samples for the $i^{th}$ motion category; N is the batch size; $C_1$ and $C_2$ are used to balance binary and STA loss. We choose $C_1=1$ and $C_2=0.5$ in this paper and train our network using Adam optimizer \cite{adam} with 0.001 initial learning rate. The training process converges fast with batch size 40 (<5,000 iterations).

\section{Experiments}

\subsection{Dataset}\label{dataset}
We collect 38,360 video clips from 78 home surveillance cameras. Each video is 15s and captured with FPS 10 and 1280$\times$720 resolutions. The videos cover different times and various scenes, e.g., front door, backyard, street and indoor living room. The longest period a camera recorded is around 3 days. Those videos mostly capture only stationary background or irrelevant motion caused by shadows, lighting changes or snow/rain. Some of the videos contain relevant motion caused by people and vehicles (car, bus and truck).

The ``relevant motion" in our system is defined with a list of relevant objects. We consider three kinds of relevant motion: ``People motion", caused by object ``people"; ``Vehicle motion", caused by at least one object from \{car, bus, truck\}; ``P+V Motion" (P+V), caused by at least one object from \{people, car, bus, truck\}. The detection performance of ``P+V Motion" evaluates the ability of our method to detect general motion, and the detection performance of ``People/Vehicle motion" evaluates the ability of differentiating motion caused by different kinds of objects. 

Our network is trained using weak supervisions obtained from the object detection based method. We run Faster R-CNN based object detection with FPS 10 on the original 1280$\times$720 video frames, and apply a state-of-the-art real-time online tracker from \cite{sort} to capture temporal consistency to obtain labels of relevant motion involving different objects-of-interest.
We first run the above method on the entire dataset to detect videos with relevant motion of P+V. Then, we randomly split the dataset into training and testing set with a 3:1 ratio. This leads to a training set with 28,732 video clips and a testing set with 9,628 video clips. 

We have three annotators watching the testing videos and annotating binary labels for P+V, people and vehicle motion. If there is any human recognizable motion of the relevant objects in a video, we annotate ``1" for the corresponding category. All three annotators must agree with each other, otherwise we mark the video as ambiguous and remove it from the testing set. We have 9,606 testing videos left after several rounds of annotation and cross-checking. Among the testing set, there are 973 videos with relevant motion events contain either people or vehicles (P+V), 429 videos have people motion only, 594 videos have vehicle motion only and 50 videos have both motion. 

\subsection{Baseline: Object Detection based Method}
For the object detection based method, if one tracklet has at least two frames overlapped with the detected bounding boxes of the relevant objects with Intersection over Union (IOU) $>$ 0.9, the average detection confidence score $>$ 0.8, and the maximum relative location change ratio over the entire tracklet along width or height of the frame is large enough ($>$ 0.2), we consider it as a valid tracklet. If there is at least one valid tracklet of an object in a video clip, we consider that video has valid motion of the specific object. 

For background subtraction, we utilize the method proposed in \cite{background}. We employ the start-of-the-art real-time online tracking method from \cite{sort}. The above two methods can be efficiently run on CPU. 
For object detection, we utilized the Tensorflow Object Detection API \cite{API}. Following the discussions in \cite{API}, we choose three state-of-the-art detectors: Faster R-CNN \cite{faster} with ResNet 101 \cite{resnet}, SSD \cite{SSD} with inception V2 \cite{inception} and SSD with MobileNet \cite{mobile}. 
According to \cite{API}, Faster R-CNN is the best detector considering detection accuracy and robustness, and is widely used in many applications \cite{rich,ang,VRD,VRD1,CWSL,zoom}. However, the SSD based detector is much faster while suffering some accuracy degradation. To further reduce the computational cost of detection, MobileNet \cite{mobile} is utilized as the base network in SSD framework. SSD-MobileNet is one of the most compact and fastest detectors.
We do not compare to YOLO v2 due to its similar performance with SSD \cite{yolo2}. We do not compare to other detectors such as tinyYOLO \cite{YOLO} and SqueezeDet \cite{squeezedet} due to their significant degradation of performance (e.g., >20$\%$ reduction in mAP from YOLO v2 to Tiny YOLO). 
We use F-score to jointly evaluate the detection precision and recall. The results are shown in Table \ref{Det_FPS}. 

\begin{table}[]
\centering
\scriptsize
\caption{F-score of relevant motion detection using different object detectors with different FPS and resolution settings. We conclude that Faster R-CNN based method significantly outperforms the SSD based ones. Meanwhile, high resolution and large FPS are needed for the object detection based method.}
\label{Det_FPS}
\begin{tabular}{c|ccccc}
\hline
                         & Detector& \begin{tabular}[c]{@{}c@{}}FPS 1\\ 1280$\times$720\end{tabular} & \begin{tabular}[c]{@{}c@{}}FPS 2\\ 1280$\times$720\end{tabular} & \begin{tabular}[c]{@{}c@{}}FPS 5\\ 1280$\times$720\end{tabular} & \begin{tabular}[c]{@{}c@{}} FPS 10\\ 1280$\times$720\end{tabular}\\ \hline
\multirow{3}{*}{P+V}  & F R-CNN    & 0.075 & 0.405 & 0.745 & \textbf{0.785}  \\
                         & SSD-Incep  & 0.103 & 0.191 & 0.403 & 0.258  \\
                         & SSD-Mobile & 0.048 & 0105  & 0.277 & 0.258  \\\hline
\multirow{3}{*}{People}  & F R-CNN    & 0.116 & 0.519 & 0.766 & \textbf{0.795}  \\
                         & SSD-Incep  & 0.156 & 0.276 & 0.467 & 0.549  \\
                         & SSD-Mobile & 0.054 & 0.139 & 0.252 & 0.258  \\\hline
\multirow{3}{*}{Vehicle} & F R-CNN    & 0.020 & 0.244 & 0.627 & \textbf{0.665}  \\
                         & SSD-Incep  & 0.062 & 0.103 & 0.267 & 0.499  \\
                         & SSD-Mobile & 0.029 & 0.062 & 0.197 & 0.252  \\ \cline{1-6} 
\hline
\hline
        & Detector & \begin{tabular}[c]{@{}c@{}}FPS 10\\ 320$\times$180\end{tabular} & \begin{tabular}[c]{@{}c@{}}FPS 10\\ 160$\times$90\end{tabular} & \begin{tabular}[c]{@{}c@{}}FPS 5\\ 320$\times$180\end{tabular} & \begin{tabular}[c]{@{}c@{}}FPS 5\\ 160$\times$90\end{tabular} \\ \hline
P+V  & F R-CNN  & \textbf{0.667}                                                 & 0.255                                                   & 0.565                                                   & 0.255                                                  \\
People  & F R-CNN  & \textbf{0.670}                                                    & 0.251                                                   & 0.574                                                   & 0.256                                            \\
Vehicle & F R-CNN  & \textbf{0.610}                                                    & 0.226                                                   & 0.517                                                   & 0.122              \\\bottomrule

\end{tabular}
\end{table}



\begin{table*}[!t]
\centering
\footnotesize
\caption{\textbf{The path from traditional 3D ConvNets to ReMotENet using Spatial-temporal Attention Model}. ``FD" denotes frame differencing; ``D" denotes separating spatial and temporal max pooling and making the network deeper; 
``MT" denotes we use the detection pseudo-groundtruth to train the STA layer to localize moving objects using multi-task learning \cite{VITON,MT}, but do not use the attention mask to scale the features extracted by Pool5; 
``T" and ``NT" denote using the attention mask predicted by STA layer to scale the features, with/without using the detection pseudo-groundtruth to train the STA layer; ``H" means using higher input resolution; ``32" denotes using more filters per layer.
There are two significant performance improvements along the path. The first is from C3D to FD-C3D by using frame differencing; The second is from FD-D to FD-D-STA-T with the spatial-temporal attention. Other design choices, e.g., larger input resolution (FD-D-STA-T-H: from 160$\times$90 to 320$\times$180) and more filters per layer (FD-D-STA-T-32: from 16 to 32) lead to comparable performance.
train STA layer with detection pseudo-groundtruth}
\label{eval}
\scriptsize
\begin{tabular}{@{}ccc|cccc|ccc@{}}
\toprule
   \begin{tabular}[c]{@{}c@{}}Network structures\\ of ReMotENet\end{tabular}    & C3D         & FD-C3D   & FD-D     & FD-D-MT  & \begin{tabular}[c]{@{}c@{}}FD-D\\ -STA-NT\end{tabular} & \begin{tabular}[c]{@{}c@{}}FD-D\\-STA-T\end{tabular} & \begin{tabular}[c]{@{}c@{}}FD-D-\\STA-T-H\end{tabular} & \begin{tabular}[c]{@{}c@{}}FD-D-\\STA-T-32\end{tabular} & \begin{tabular}[c]{@{}c@{}}FD-D-STA\\-T-H-32\end{tabular} \\ \midrule
3D ConvNets?                 & \checkmark & \checkmark & \checkmark & \checkmark & \checkmark                                               & \checkmark                                              & \checkmark                                                & \checkmark                                                 & \checkmark                                                   \\
frame differencing?                                   &            & \checkmark & \checkmark & \checkmark & \checkmark                                               & \checkmark                                              & \checkmark                                                & \checkmark                                                 & \checkmark                                                   \\
deeper network?                        &            &            & \checkmark & \checkmark & \checkmark                                               & \checkmark                                              & \checkmark                                                & \checkmark                                                 & \checkmark                                                   \\
train STA layer with detection pseudo-GT?                   &            &            &            & \checkmark &                                                          & \checkmark                                              & \checkmark                                                & \checkmark                                                 & \checkmark                                                   \\
ST Attention? &                        &            &            &            & \checkmark                                               & \checkmark                                              & \checkmark                                                & \checkmark                                                 & \checkmark                                                   \\
high resolution?                 &            &            &            &            &                                                          &                                                         & \checkmark                                                &                                                            & \checkmark                                                   \\
more filters?     &            &                        &            &            &                                                          &                                                         &                                                           & \checkmark                                                 & \checkmark                                                   \\ \midrule
AP: P+V                  & 77.79         & 82.29      & 83.98      & 84.25      & 84.91                                                    & 86.71                                                   & 85.67                                                     & \textbf{87.07}                                                      & 86.09                                                        \\
AP: People                 & 62.25         & 72.21      & 73.69      & 74.41      & 75.82                                                    & 78.95                                                   & \textbf{79.78}                                                     & 77.92                                                      & 77.54                                                        \\
AP: Vehicle                 & 66.13          & 73.03      & 73.71      & 74.25      & 75.47                                                    & \textbf{77.84}                                                  & 76.85                                                     & 76.81                                                      & 76.92                                                        \\ \bottomrule
\end{tabular}
\end{table*}

From the tables we conclude that Faster R-CNN based detector achieves much better performance than the other two efficient detection methods. Meanwhile, image resolution and frame sample rate (FPS) have significant influence. If resolution
or FPS is small,
the performance of the detector and tracker drops significantly. So, to achieve reasonable detection results, we need to employ a robust object detection framework (Faster R-CNN) with large FPS and resolutions, which is inefficient and heavy (with model size $>$ 400MB). 

We also conduct experiments without tracking. In general, tracking improves precision but leads to lower recall. Taking the detection of P+V motion using Faster R-CNN with FPS 10 and 1280$\times$720 resolution as an example, 
without tracking, the method will have high recall (e.g., 0.8536) but very low precision (e.g., 0.2631); with tracking, it has lower recall (e.g., 0.7321) but much higher precision (e.g., 0.8467), and achieves much higher F-score.
The detailed results of different settings of the object detection based method are included in the supplementary materials.

\subsection{ReMotENet Performance}
The outputs of ReMotENet are three binary predictions. After applying softmax on each binary prediction, we obtain probabilities of having P+V motion, people motion and vehicle motion in a video clip. We adopt Average Precision, which is a widely used evaluation metric for object detection and other detection tasks from \cite{API} to evaluate our method. 
We evaluate different architectures and design choices of our methods, and report the average precision of detecting P+V motion, people motion and vehicle motion in Table \ref{eval}. 
To show the improvement of our proposed method, we build a basic 3D ConvNets following \cite{C3D}. We design a 3D ConvNets with 5 Conv layers followed by spatial-temporal max pooling. Similar to the C3D in \cite{C3D}, we conduct 3$\times$3$\times$3 3D convolution with 1$\times$1$\times$1 stride for Conv1-Conv5, and 2$\times$2$\times$2 spatial-temporal max pooling with 2$\times$2$\times$2 stride on Pool2-Pool5. For Pool1, we conduct 1$\times$2$\times$2 spatial max pooling with 1$\times$2$\times$2 stride. Different from C3D in \cite{C3D}, we only have one layer of convolution in Conv1-Conv5. Meanwhile, instead of several fully connected layers, we apply a global average pooling followed by several 1$\times$1$\times$1 convolution layers after Conv5.
The above basic architecture is called ``C3D" in Table \ref{eval}. 

\begin{figure}[!t]
\centering
  \includegraphics[height=4cm]{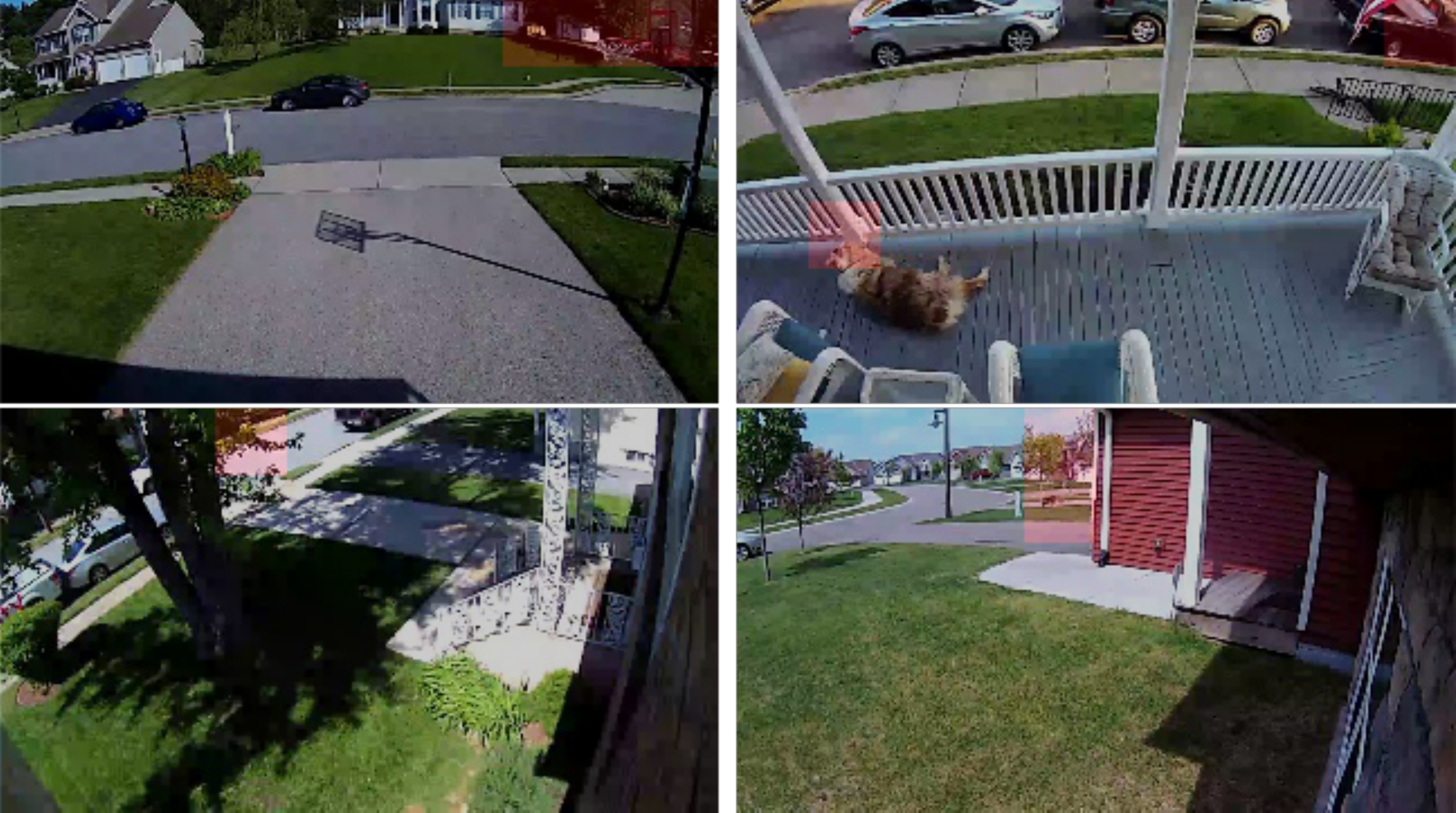}
  \caption{Predicted Attention Mask of ``FD-D-STA-NT" Method. Without pseudo-groundtruth bounding boxes of the semantic moving relevant objects obtained from the object detection based method, the attention model will focus on some ``irrelevant" motion caused by the objects outside the pre-specified relevant object list, e.g., pets, trees and flags. The overlaid red rectangles are the predicted motion masks (has probability $>$ 0.9).}
\label{fig:NoT}
\end{figure}

\begin{figure*}[!t]
\centering     
\subfigure[PR Curve: P+V]{\label{fig:PR_M}\includegraphics[width=.25\linewidth]{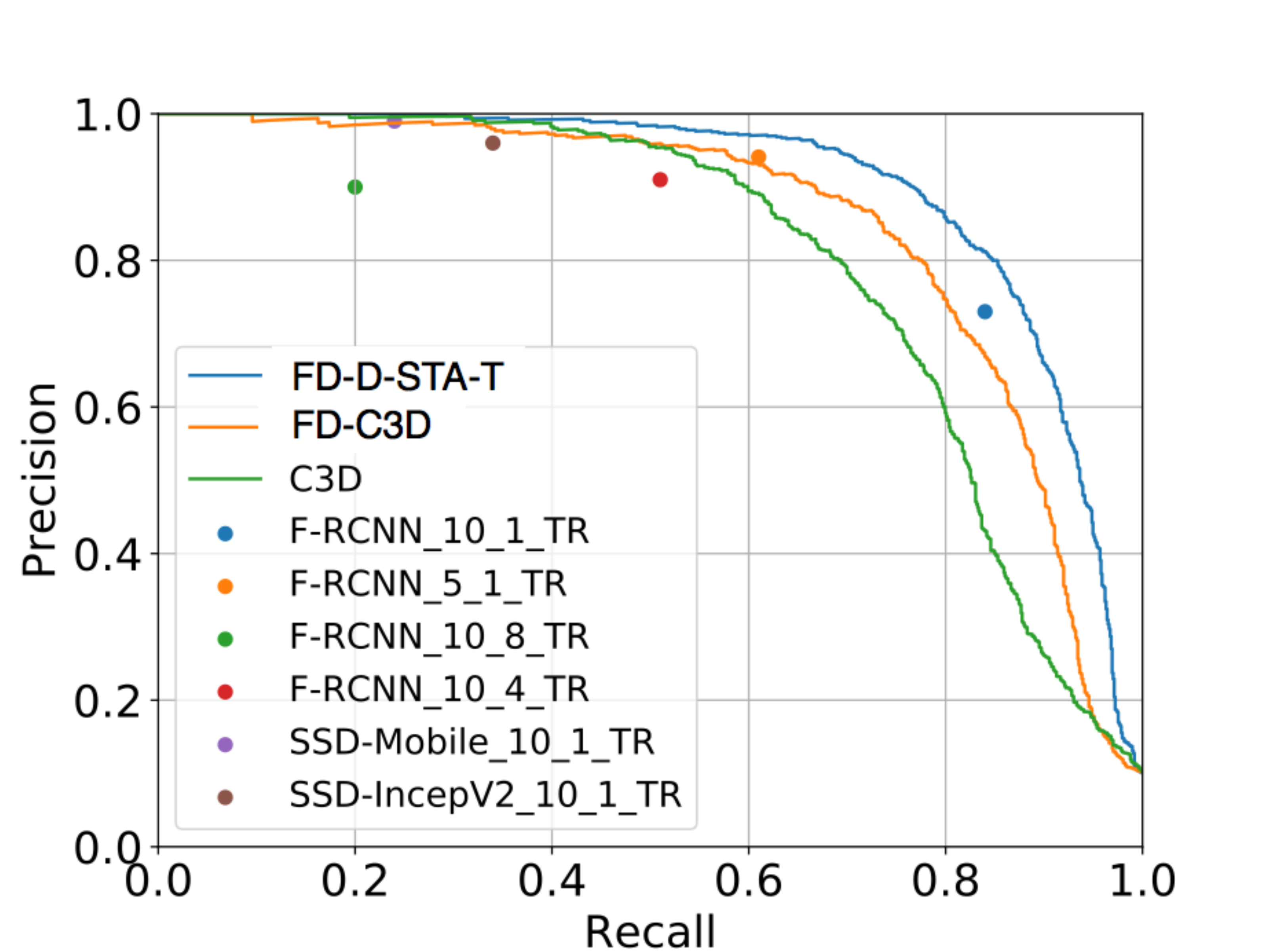}}
\subfigure[PR Curve: People]{\label{fig:PR_P}\includegraphics[width=.25\linewidth]{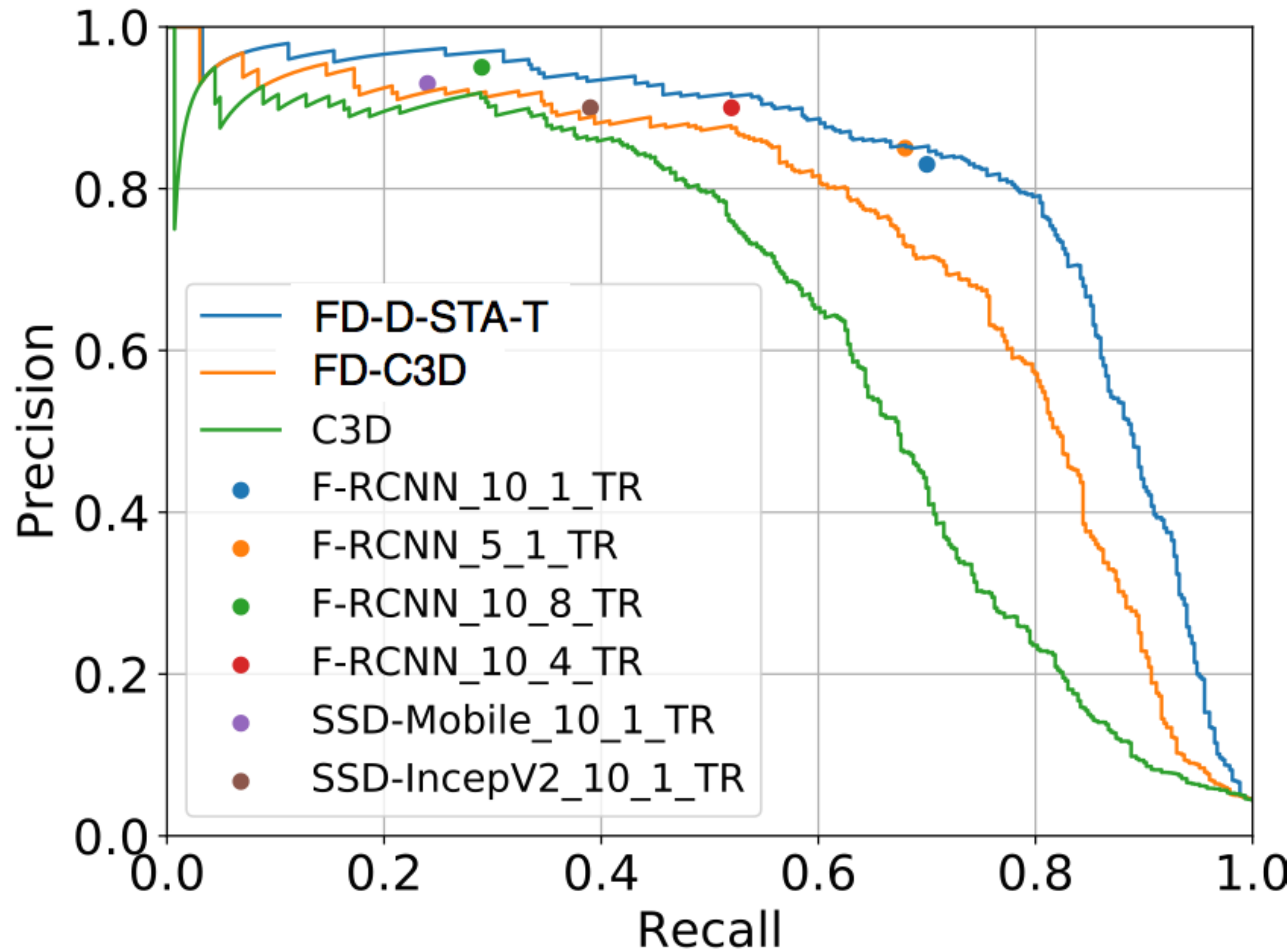}}
\subfigure[PR Curve: Vehicle]{\label{fig:PR_V}\includegraphics[width=.25\linewidth]{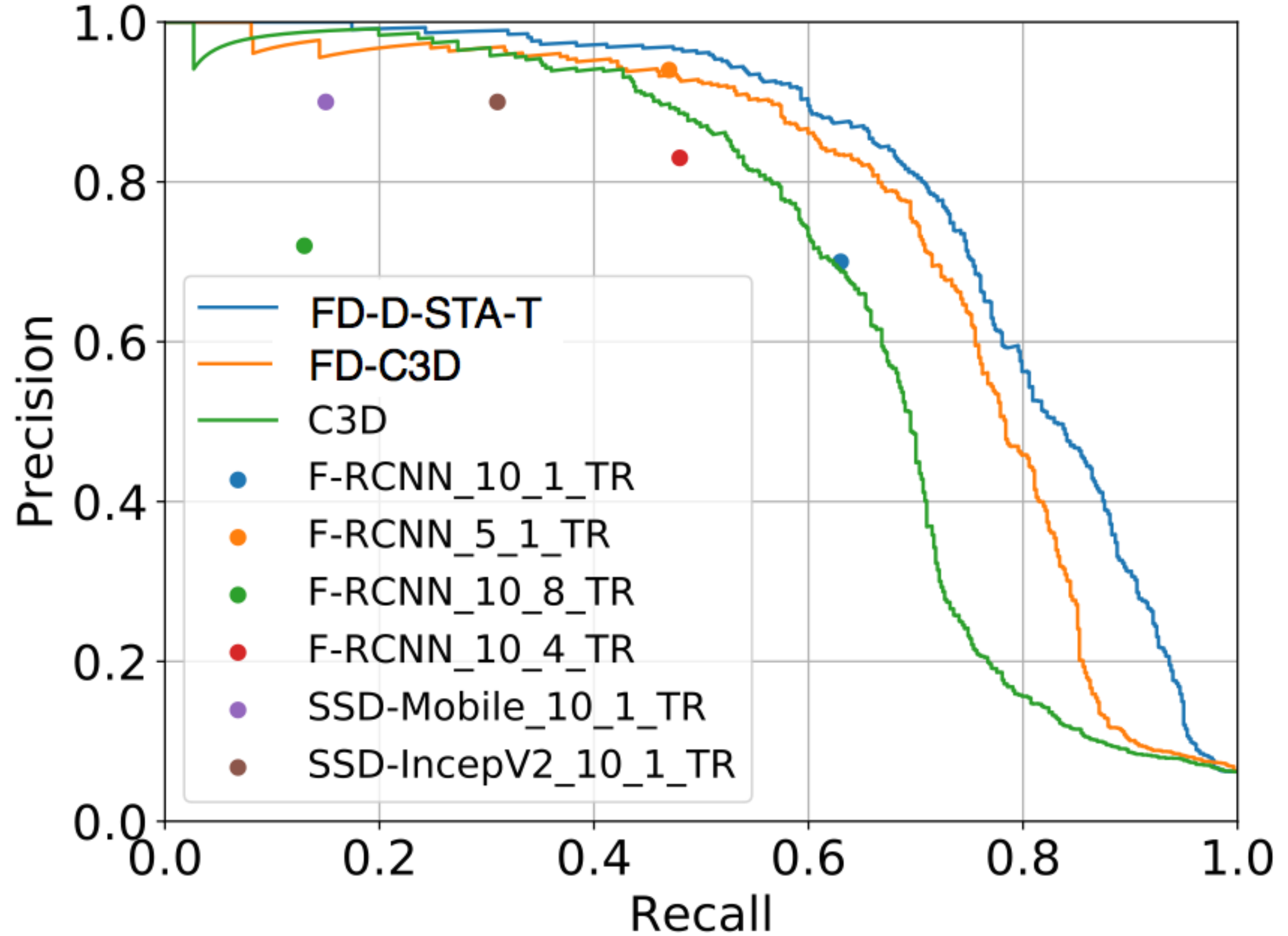}}

\subfigure[Run-time per 15s video on GPU (second)]{\label{fig:GPU}\includegraphics[width=.25\linewidth]{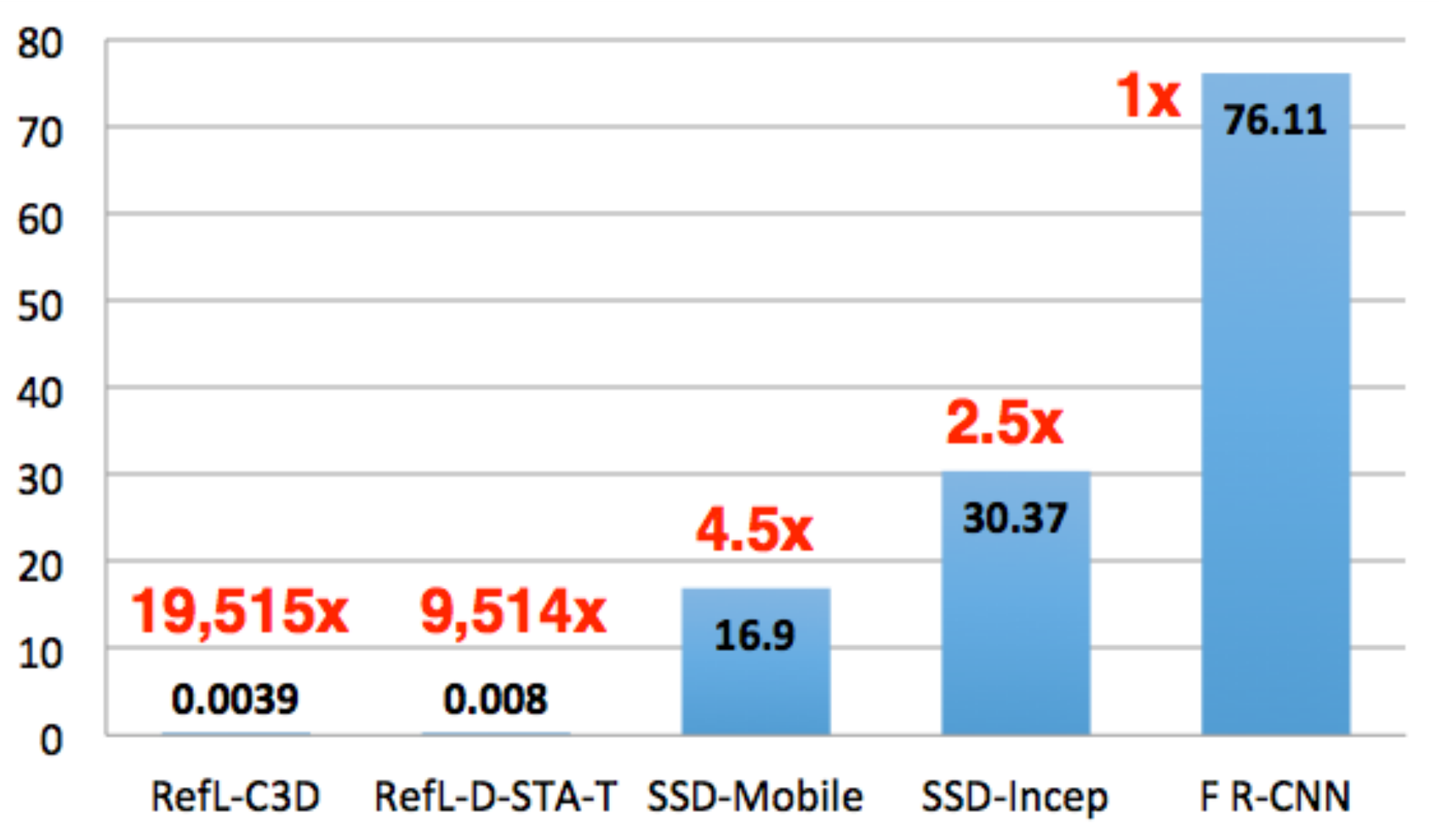}}
\subfigure[Run-time per 15s video on CPU (second)]{\label{fig:CPU}\includegraphics[width=.25\linewidth]{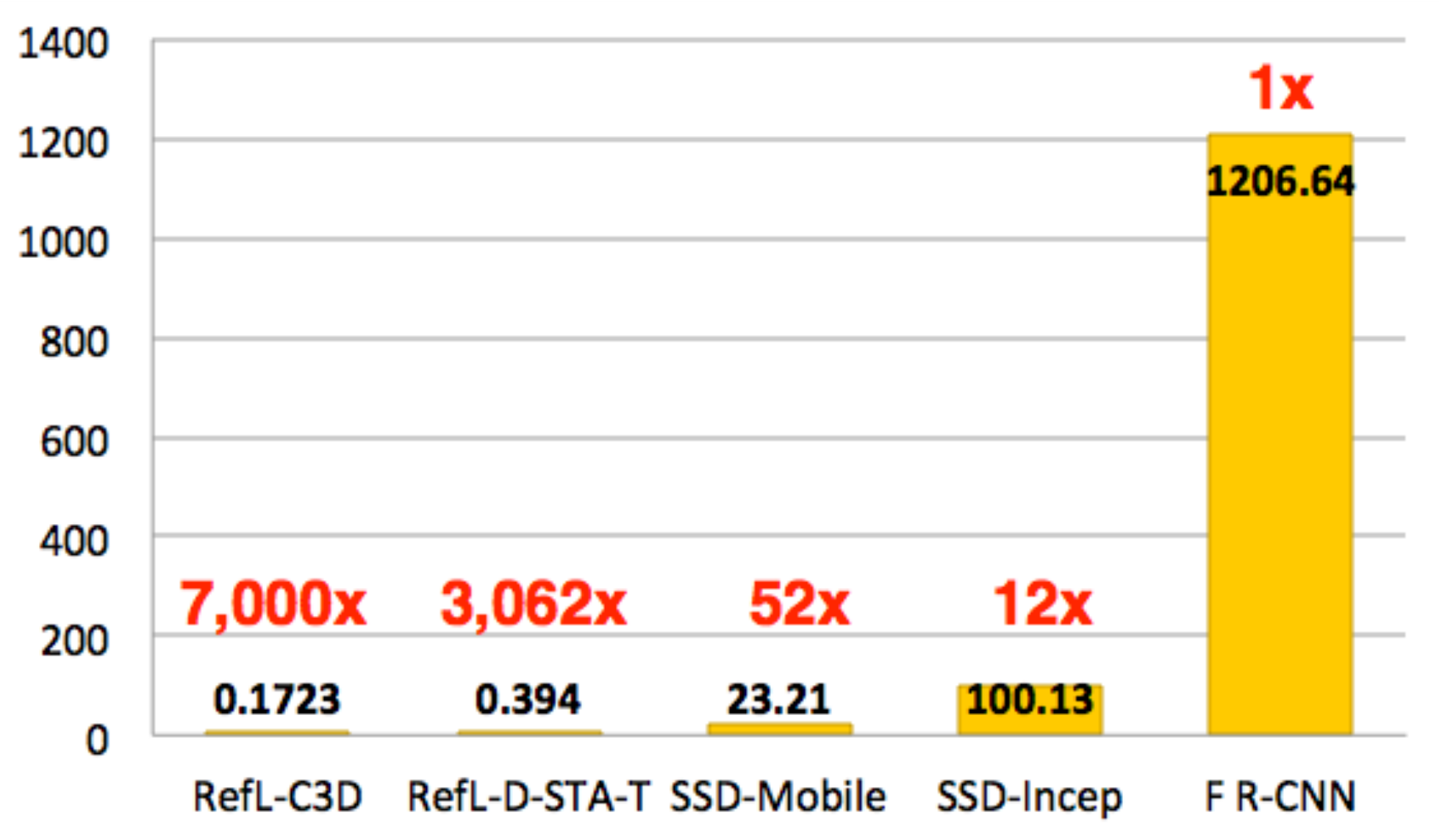}}
\subfigure[Model Size (KB)]{\label{fig:size}\includegraphics[width=.25\linewidth]{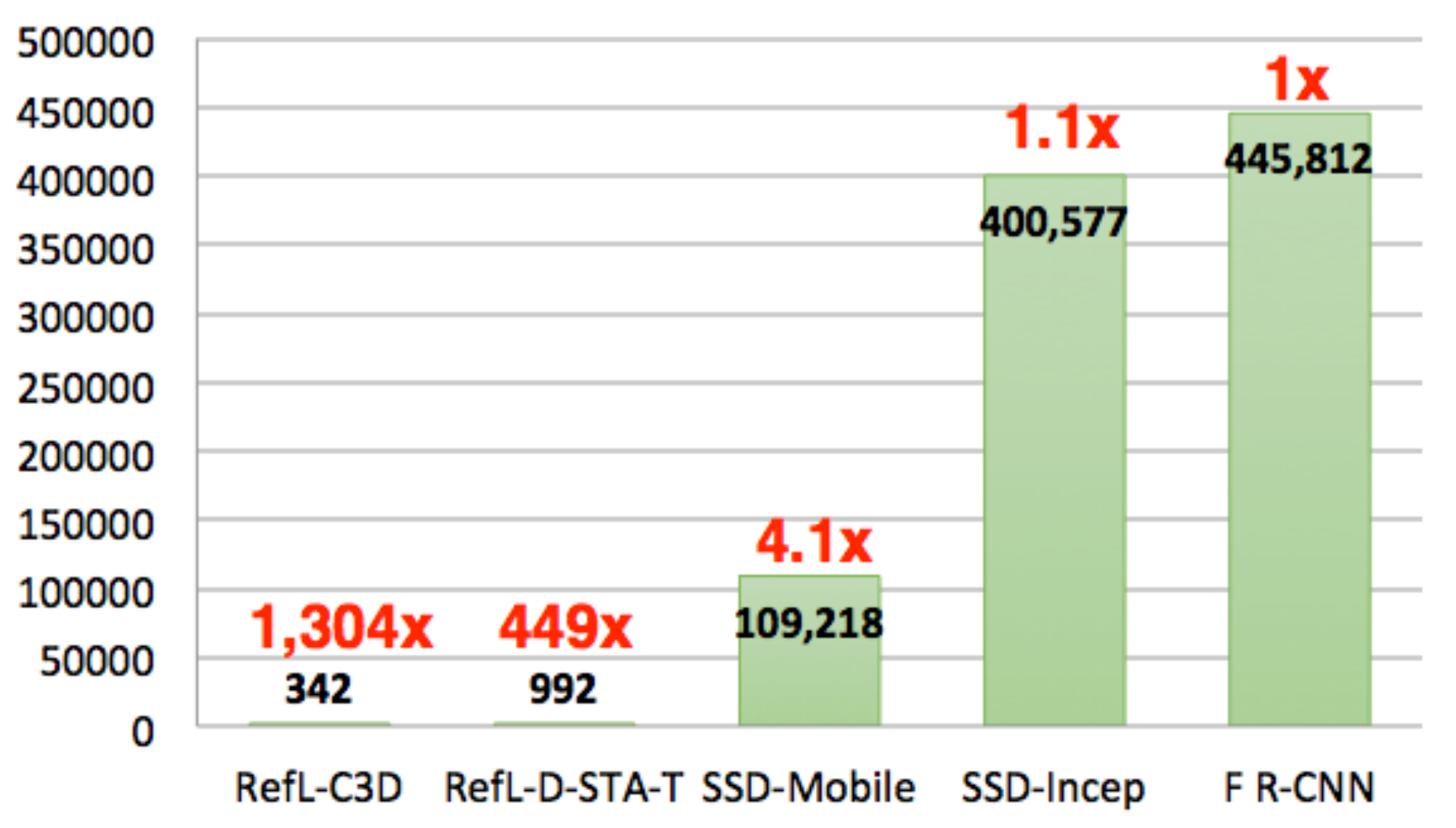}}
\caption{Comparing with baseline. (a)-(c) show the precision-recall (PR) curves of our methods and the precision-recall of object detection based methods. For all three motion categories, the PR curve of our ``FD-D-STA-T" method is always higher than all the PR points of object detection based methods, which means by properly setting the probabilistic threshold, our method can achieve comparable or better detection performance than the baselines. 
The notation of object detection based methods is constructed as ``detector\_FPS\_resolution reduction factor\_tracking" 
(d)-(f) show the run-time and model size comparison between our methods and the baselines. 
}
\end{figure*}

\subsubsection{Evaluation of ReMotENet}
First, we evaluate the effect of frame differencing. Column 2-3 in Table \ref{eval} show that by using frame differencing (FD), our 3D ConvNets achieve much higher average precision for all three categories of motion, especially on people and vehicle motion detection tasks. 

To evaluate the spatial-temporal attention model, we modify the basic C3D network architecture to separate spatial and temporal max pooling as shown in Table \ref{network}. We call the network with nine 3D ConvNets (without the STA layer) ``FD-D". ``FD-D-MT" denotes using multi-task learning: we use the STA layer to predict the ST attention mask, which is trained by pseudo-groundtruth obtained from the object detection based method, but we do not multiply the attention mask with the extracted features after Pool5. Another model is ``FD-D-STA-NT": we multiply the attention mask with the extracted features after Pool5 layer. However, the STA layer is trained with only binary labels of motion categories but without the detection pseudo-groundtruth. 
We observe that incorporating multi-task learning or end-to-end attention model only leads to a small improvement, but once we combine both methods, our ``FD-D-STA-T" model achieves significant improvement. Meanwhile, due to the sparsity of moving relevant objects in the videos, the number of positive and negative spatial-temporal location from the detection pseudo-groundtruth is extremely biased. This leads to overfitting of the model to predict the attention of all the spatial-temporal locations as 0. 
Meanwhile, adding the attention model without multi-task learning also leads to small improvement. We observe that without the weak supervision of specific objects and their locations, the attention mask predicted by ``FD-D-STA-NT" may focus on motion caused by irrelevant objects, such as pets, trees and flags shown in Figure \ref{fig:NoT}.

To encourage our network to pay attention to the relevant objects (in this paper, people and vehicles), we propose our full model ``FD-D-STA-T", which can be viewed as a combination of multi-task learning and attention model. We use the detected bounding boxes of relevant moving objects to train the STA layer, and multiply the predicted attention mask with the extracted features from Pool5 layer. 
``FD-D-STA-T" achieves much higher average precision than the previous models in all three motion categories. 
The results of the above models are listed in column 5-8 of Table \ref{eval}. 

We also conduct experiments with other network design choices of ReMotENet. For instance, we add more filters in each convolution layer, or enlarge the input resolution from 160$\times$90 to 320$\times$180. As shown in Table \ref{eval}, those design choices lead to insignificant improvements. The experiments demonstrate that ReMotENet can precisely detect relevant motion with small input FPS and resolution.

\subsubsection{Comparing with the Object Detection based Method}
Although ReMotENet is trained with the pseudo-groundtruth obtained from the object detection based method, it outperforms the baseline in several ways:

\textbf{Detection Performance}. Although the training labels are noisy, ReMotENet can learn patterns of relevant motion and generalize well. We show the PR curve of our methods and the performance of the object detection based method in Figure \ref{fig:PR_M}, \ref{fig:PR_P} and \ref{fig:PR_V}. From the PR curve of three models: ``C3D", ``FD-C3D" and ``FD-D-STA-T", we demonstrate the effectiveness of frame differencing and the STA model. The PR curves of our full model ``FD-D-STA-T" are higher than all PR points of the object detection based method, which shows that our method can achieve similar or better performance than the baselines by properly choosing the detection threshold. Another advantage of our method is that it is a probabilistic model, and one can tune the threshold based on the priority of precision or recall.
  
\textbf{Run-time and Model Size}. From Table \ref{Det_FPS} we observe that with a fixed detector, the object detection based method needs large FPS and resolutions to achieve good detection performance, especially for SSD based detectors. So, we show the time and model size benchmark results of the baselines with FPS 10 and 1280$\times$720 resolution, which achieves the best detection performance, in Figure \ref{fig:GPU}, \ref{fig:CPU} and \ref{fig:size}. 
For baselines, the run-time consists of the time for background subtraction, object detection and tracking; for ReMotENet, the run-time consists of frame differencing and the forward pass of our 3D ConvNets. We omit the time for decoding and sampling frames from the video clips for both methods. The model size is the size of Tensorflow model data file. We omit the size of meta and index file.
Our method can achieve a 9,514$\times$-19,515$\times$ speedup on GPU (GeForce GTX 1080) and a 3,062$\times$-7,000$\times$ speedup on a CPU (Intel Xeon E5-2650 @2.00GHz). Meanwhile, since our 3D ConvNet is fully-covolutional, and very compact (16 filters per Conv layer), we can achieve a 449$\times$-1,304$\times$ reduction on model size. Our ``FD-C3D" model can analyze more than 256 15s video clips per second on a GPU and around 6 videos on a CPU with a 300KB model; ``FD-D-STA-T" can parse 125 or 2.5 videos of 15s on GPU or CPU devices respectively with a model smaller than 1MB. Our method not only efficiently analyze home surveillance videos on cloud with GPUs, but can also be potentially run on the edge. 

\subsection{Visualization}

\begin{figure}
\centering     
\subfigure[Failure Cases: Object Detection based method]{\label{fig:Det-fail}\includegraphics[width=1\linewidth]{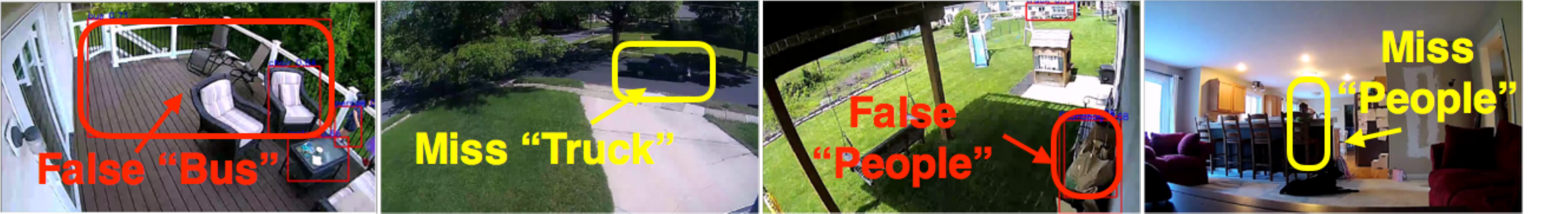}}

\subfigure[Spatial-temporal Attention Prediction: ReMotENet (FD-D-STA-T)]{\label{fig:STA}\includegraphics[width=1\linewidth]{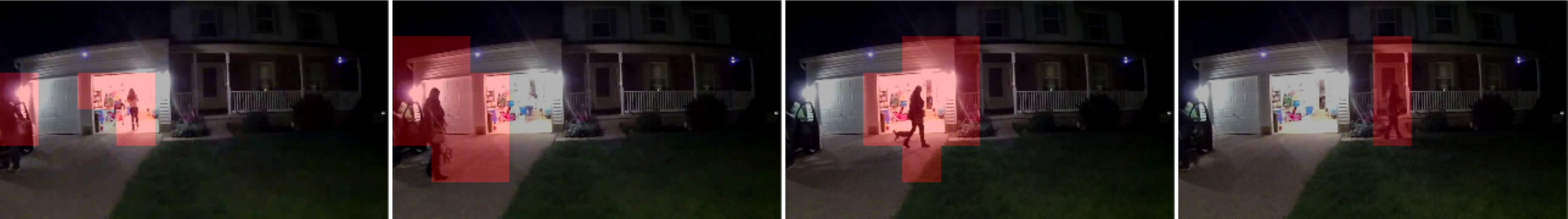}}

\subfigure[Spatial-temporal Attention: Pseudo-groundtruth]{\label{fig:STA-GT}\includegraphics[width=1\linewidth]{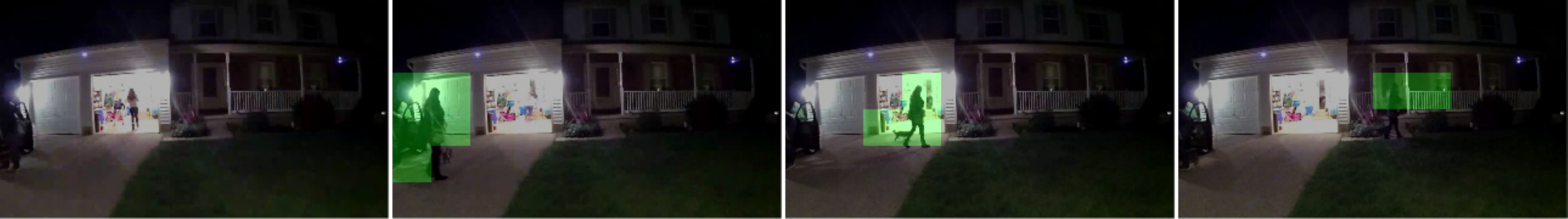}}

\subfigure[Failure Cases: ReMotENet (FD-D-STA-T)]{\label{fig:STA-fail}\includegraphics[width=1\linewidth]{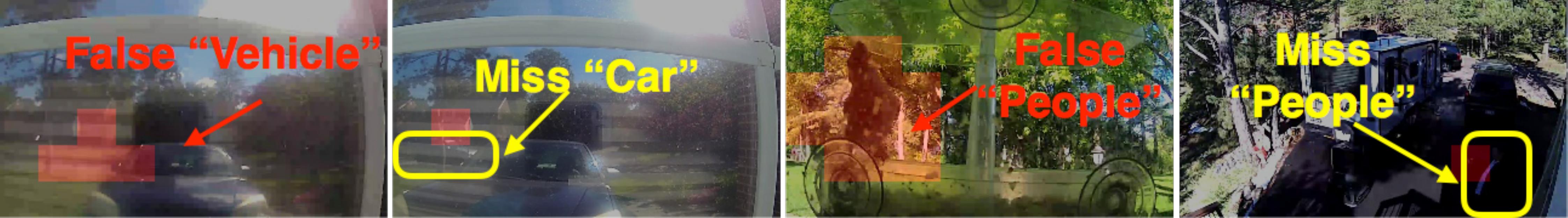}}
\caption{Visualization of results from the detection based method and ReMotENet. The details about how to generate pseudo-groundtruth in (c) can be found in the supplementary materials.}
\end{figure}

We visualize the results of the baseline and our method in Figure \ref{fig:Det-fail} to \ref{fig:STA-fail}.
Figure \ref{fig:Det-fail} shows miss/false detection of people and vehicle motion using the object detection based method. Due to the low quality of video, shadows, small object size and occlusions, object detector may fail. However, because our method is a data-driven framework trained specifically on the home surveillance video with the above properties, and it jointly models both object appearance and motion, we can handle the above cases.
Figure \ref{fig:STA} and \ref{fig:STA-GT} show the predicted spatial-temporal attention mask (red rectangles) of our ``FD-D-STA-T" model and the pseudo-groundtruth obtained from the object detection based method. Although our ``FD-D-STA-T" model can only predict a coarse attention mask, it captures the motion pattern of the people in the video. Meanwhile, although trained by the pseudo-groundtruth, our method can recover from mistakes of the object detection based method (see the first figures of \ref{fig:STA} and \ref{fig:STA-GT}). 
Figure \ref{fig:STA-fail} shows several failure cases of our method. For the miss detection (the second and forth figures), although our 3D ConvNet predicts there is no relevant motion, it still correctly predicts the coarse attention mask that covers the car and people. For the falsely detected vehicle motion (the first figure), we find that due to reflection of light, the input after frame differencing is very noisy; for the false people motion detection (the third figure), our method falsely detects bird motion as people. 

\section{Conclusion}
We propose an end-to-end data-driven framework to detect relevant motion from large-scale home surveillance videos. Instead of parsing a video in a frame-by-frame fashion using the object detection based method, we proposed to use 3D ConvNets to parse an entire video clip at once to dramatically speedup the process. 
We extended the 3D ConvNets by incorporating a spatial-temporal attention model to encourage the network to pay more attention to the moving objects. Evaluations demonstrate that ReMotENet achieves comparable or better performance than the object detection based method while achieving three to four orders of magnitude speedup on GPU and CPU devices. ReMotENet is very efficient and compact, and therefore naturally implementable on the edge (camera). 

\small{\section*{Acknowledgement}
The research was done at the Comcast Applied AI Research. Partial support from the Office of Naval Research under Grant N000141612713 (Visual Common Sense Reasoning for Multi-agent Activity Prediction and Recognition) is acknowledged. The fruitful discussion with MD Mahmudul Hasan, Upal Mahbub and Jan Neumann is highly appreciated. Special thanks go to all Comcast TPS volunteers who donated their home videos. } 

\section{Supplementary Materials}
We show more evaluation results in this material including performance evaluation of the object detection based method with different settings; different reference-frames for frame differencing; detection performance v.s. training iterations showing good converging property of training ReMotENet; the benchmark of ReMotENet with the object detection based method (with vs. without tracking); and the details of how we obtained the pseudo-groundtruth from the object detection based method.


\subsection{Precision and Recall for Object Detection based Method with Different Settings}

\begin{figure*}
\centering     
\subfigure[P+V]{\label{fig:PRD_M3}\includegraphics[width=.25\linewidth]{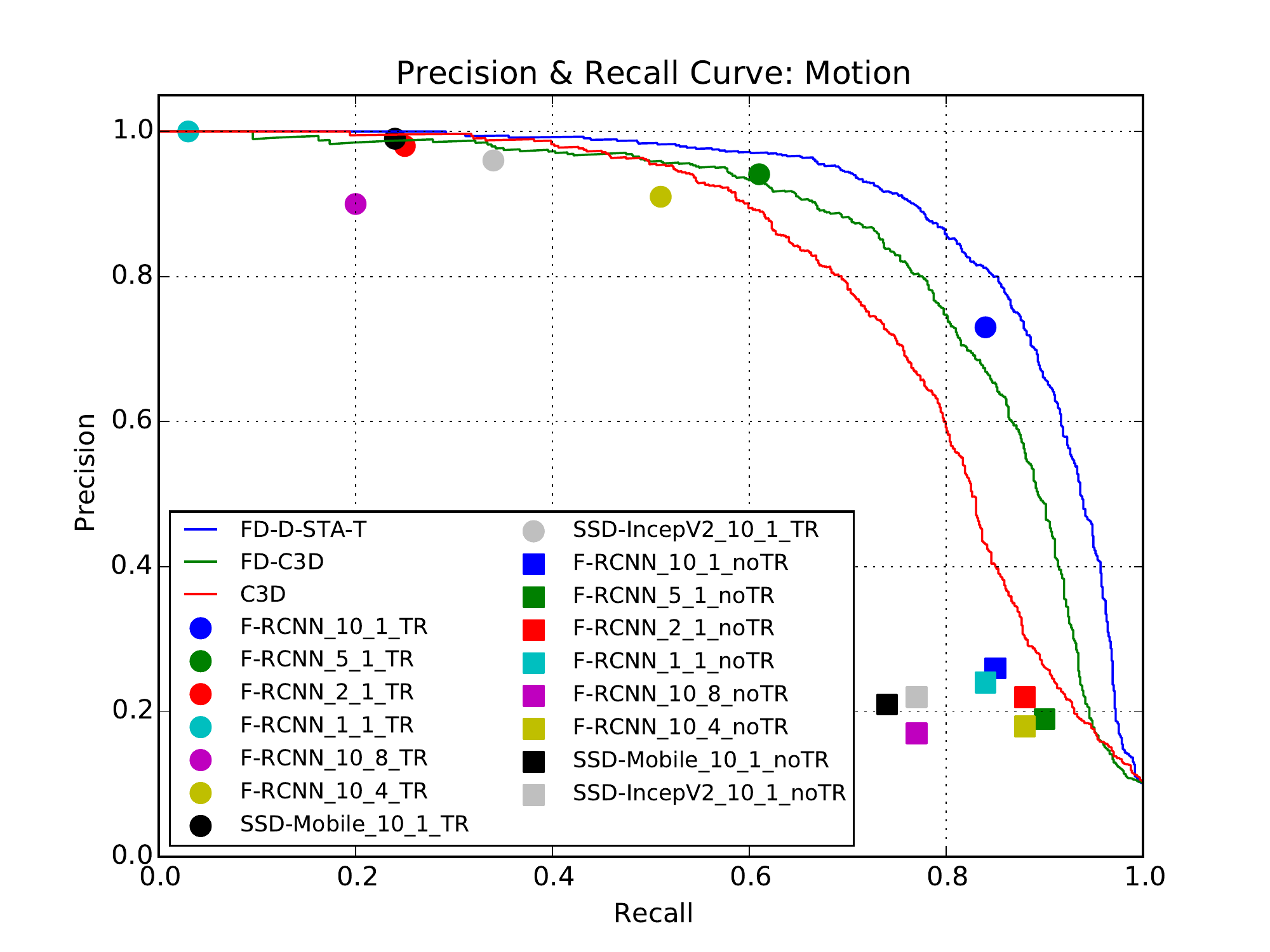}}
\subfigure[Person]{\label{fig:PRD_P3}\includegraphics[width=.25\linewidth]{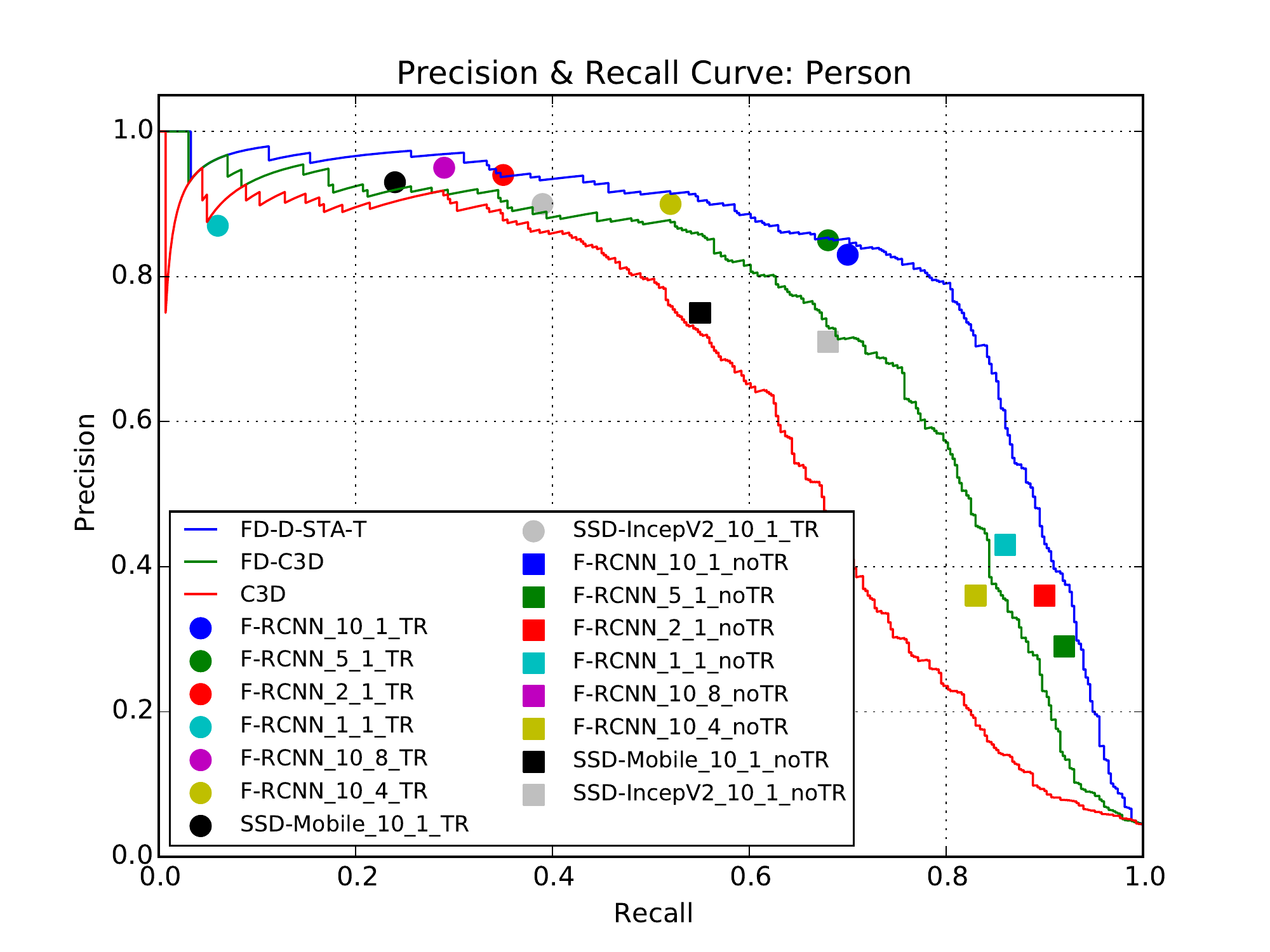}}
\subfigure[Vehicle]{\label{fig:PRD_V3}\includegraphics[width=.25\linewidth]{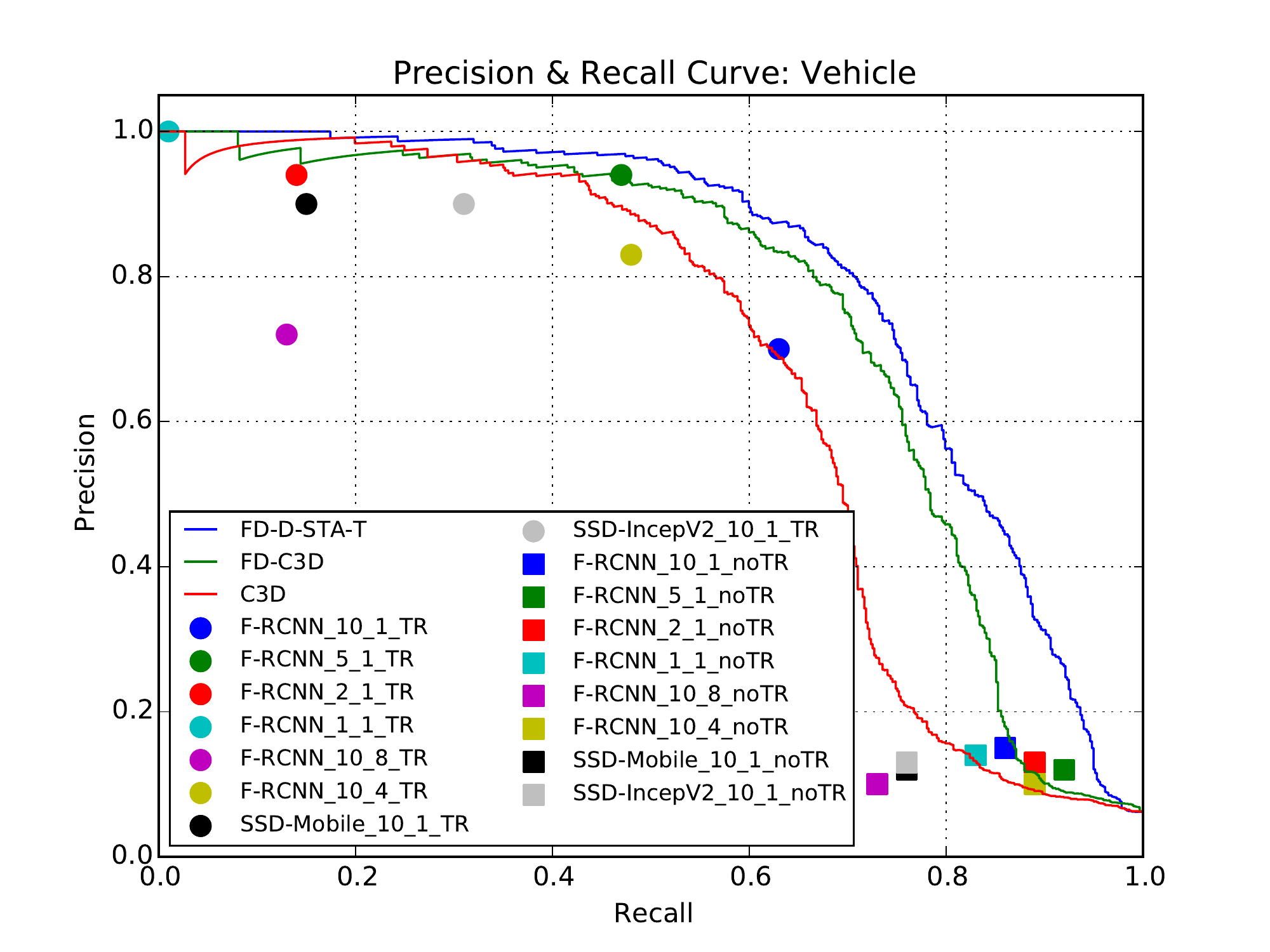}}
\caption{Precision and Recall for Object Detection based Method with Different Settings. The notation of object detection based methods is constructed as ``detector\_FPS\_resolution reduction factor\_with(out) tracking". 
The circles denote the object detection based methods with tracking; the rectangles denote without tracking.
}
\end{figure*}
In contrast to traditional sequential prediction methods \cite{lstm,rao1,rao2,rao3}, our method takes an entire video clip as input.
Figure \ref{fig:PRD_M3}-\ref{fig:PRD_V3} shows the precision and recall comparison w.r.t. the object detection based method with and without object tracking. We can see that for the object detection based method, without object tracking, a high recall can be achieved, but the precision is very low due to a large number of temporally inconsistent false detections; with object tracking, the method achieved a much higher precision with slightly lower recall; the better overall performance (F-Score) was achieved with tracking.
Meanwhile, the object detection based method requires robust and powerful but less efficient detection framework (such as Faster R-CNN), high resolution and large FPS to achieve good performance. Efficient detection framework (SSD) and compact network (MobileNet) suffer performance degradation.
As shown in Figure \ref{fig:PRD_M3}-\ref{fig:PRD_V3}, ReMotENet achieves comparable or better performance when compared to the object detection based method with or without tracking.

\subsection{Experiments: ReMotENet}

\subsubsection{Visualization of Frame Differencing}

\begin{figure}
\centering     
\subfigure[Person Motion]{\label{fig:RefL1}\includegraphics[width=.45\linewidth]{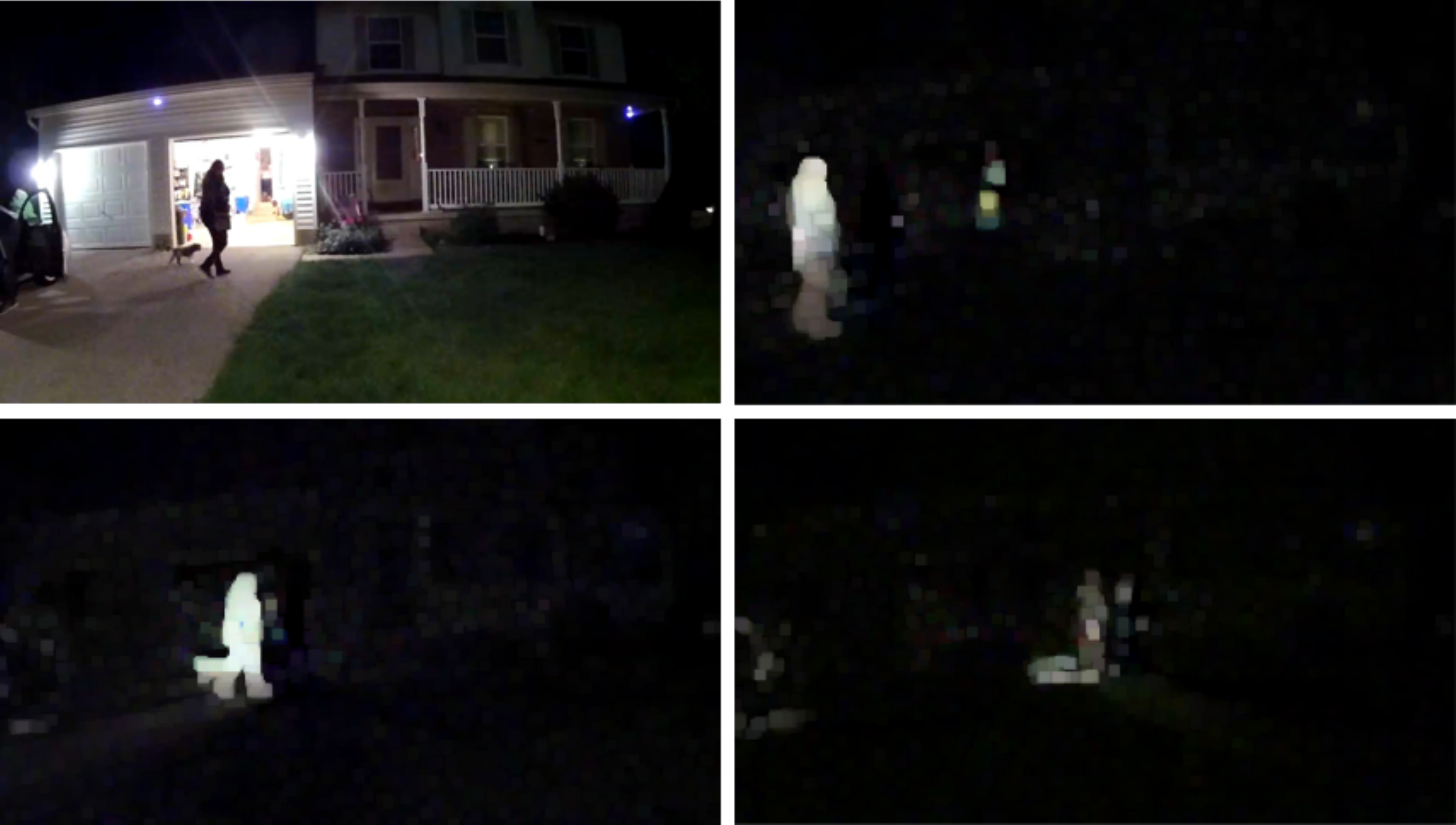}}
\subfigure[No Motion]{\label{fig:RefL2}\includegraphics[width=.45\linewidth]{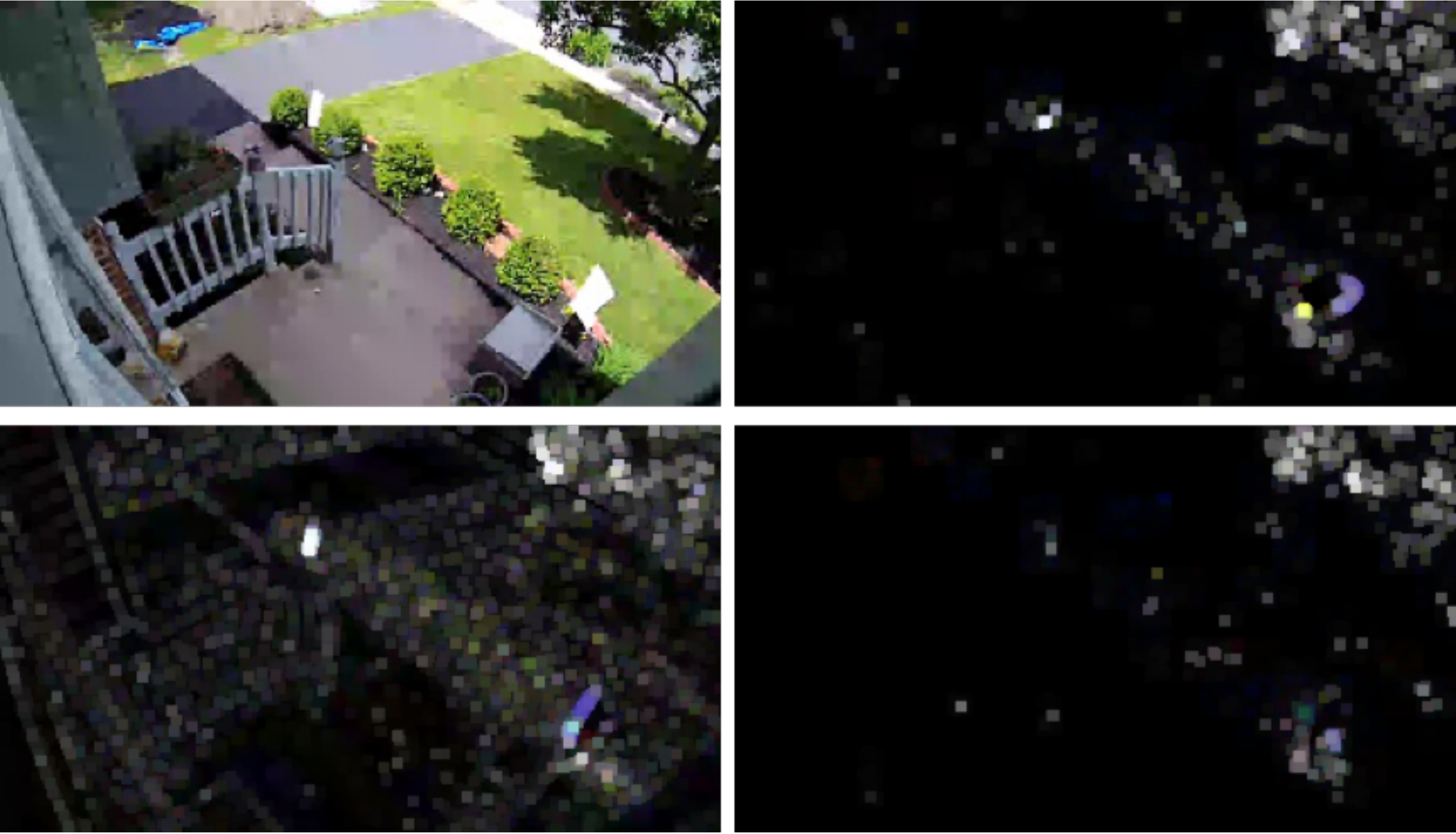}}
\caption{Frame Differencing: (a) shows the frames after frame differencing from a video clip contains person motion; (b) shows the frames from a video clip without relevant motion. For better visualization, we conduct dilation on the frames.}
\end{figure}

Figure \ref{fig:RefL1} shows some frames after frame differencing for a video clip contain relevant motion (person). 
We observe that the frame differencing successfully suppresses the background and highlights the moving foreground of person.
Figure \ref{fig:RefL2} shows the frames from a video clip without relevant motion. However, due to shadow and wind, the input is not all-zero, but the frame differencing still suppresses the stationary background to reduce the learning complexity of our method.

\subsubsection{Evaluation of Frame Differencing}
\begin{figure}[!t]
\centering
  \includegraphics[height=4cm]{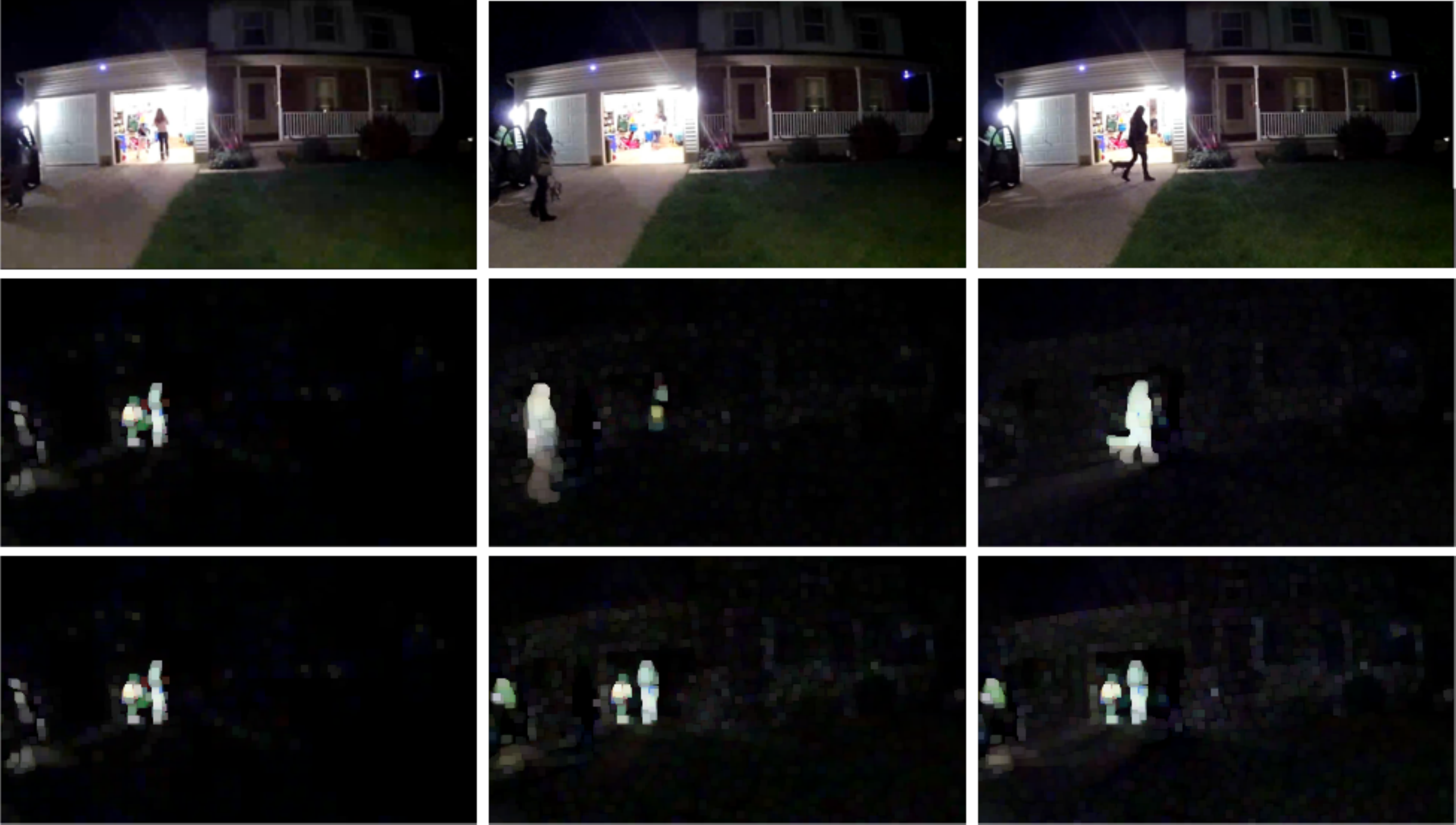}
  \caption{Comparison between different reference frames: The first row shows the raw video frames;  the second row shows frames after subtracting local reference-frame; third row shows frames after subtracting global reference-frame.}
\label{fig:refGL}
\end{figure}

First, we evaluate the effect of different frame differencing schemes in our framework. We tested two choices of reference frame: global reference-frame (RefG), which is the first sub-sampled frame of a video clip; local reference-frame (RefL), which is the previous sub-sampled frame of the current frame. We show examples of frames subtracted from RefG and RefL in Figure \ref{fig:refGL}. Our intuition is that if there are relevant objects in the first frame, and if we choose it as the global reference-frame, there will always be holes of those objects in the subsequent frames, which may be misleading for the network. Our experiments also validated that RefL leads to better performance than RefG, especially on the people and vehicle motion detection task (as shown in Fig. \ref{fig:GL}). That is why we used RefL in all the benchmark work in the paper. 

\begin{table}[!t]
\centering
\scriptsize
\caption{Evaluation of Frame Differencing (Average Precision): C3D denotes using raw frames; RefG-C3D denotes using the first frame as the reference frame; RefL-C3D denotes using the previous frame as the reference frame. Using frame differencing, especially RefL-C3D leads to significant improvement of the relevant motion detection when compared to using raw frames as input.}
\label{fig:GL}
\begin{tabular}{@{}c|ccc@{}}
\toprule
         & P+V           & P Motion      & V Motion      \\ \midrule
C3D      & 77.8          & 62.3          & 66.1          \\
RefG-C3D & 81.8          & 70.7          & 69.2          \\
RefL-C3D & \textbf{82.3} & \textbf{72.2} & \textbf{73.0} \\ \bottomrule
\end{tabular}
\end{table}

\subsubsection{Testing Performance v.s. Training Iterations}
\begin{figure}[!t]
\centering
  \includegraphics[height=4cm]{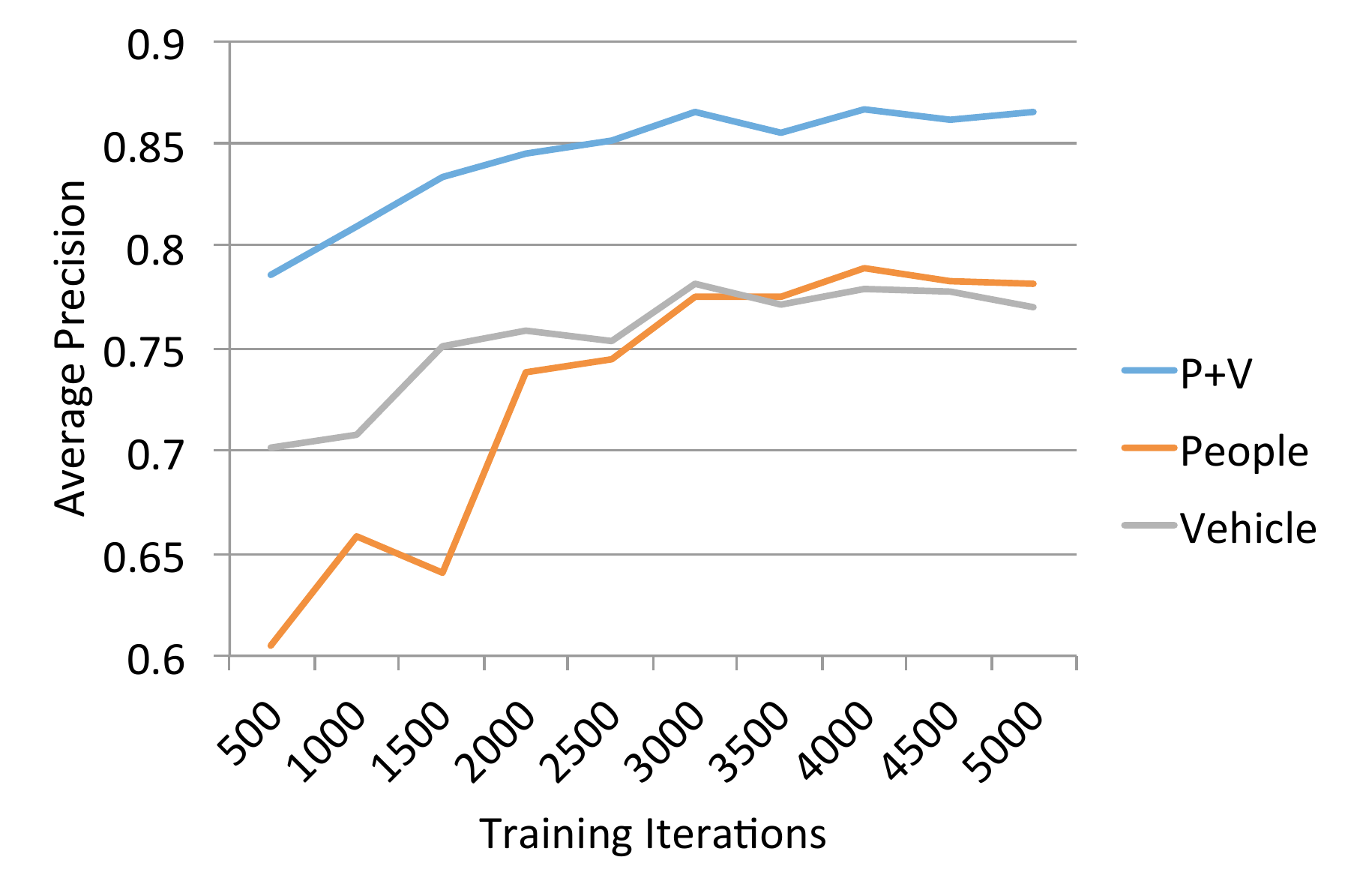}
  \caption{Testing Average Precision v.s. Number of training Iterations: RefL-D-STA-T model of ReMotENet.}
\label{fig:converge}
\end{figure}

We show the testing average precision v.s. number of training iterations in Figure \ref{fig:converge}. ReMotENet takes a small number of training iterations to achieve good detection performance. This shows good converging properties of training ReMotENets.

\subsubsection{Pseudo-groundtruth Obtained from the Object Detection based Method}
Besides the final binary labels, we also use the object detection based pseudo-groundtruth to train the STA layer and achieve significant performance improvement. To obtain the pseudo-groundtruth, we first conduct the pipeline consisting of background subtraction, object detection using Faster R-CNN with 1280$\times$720 resolution and 10 FPS and tracking, to obtain valid tracklets of the objects-of-interest in each video clip (the hyper-parameters are the same as those in section 4.2 in the main paper). Then, for each detected bounding box of a relevant object, we consider it a true positive if the detection confidence score is $>$ 0.8. For each frame, we construct a binary mask based on all detected true positive bounding boxes of the relevant objects.

\begin{figure}[!t]
\centering
  \includegraphics[height=4cm]{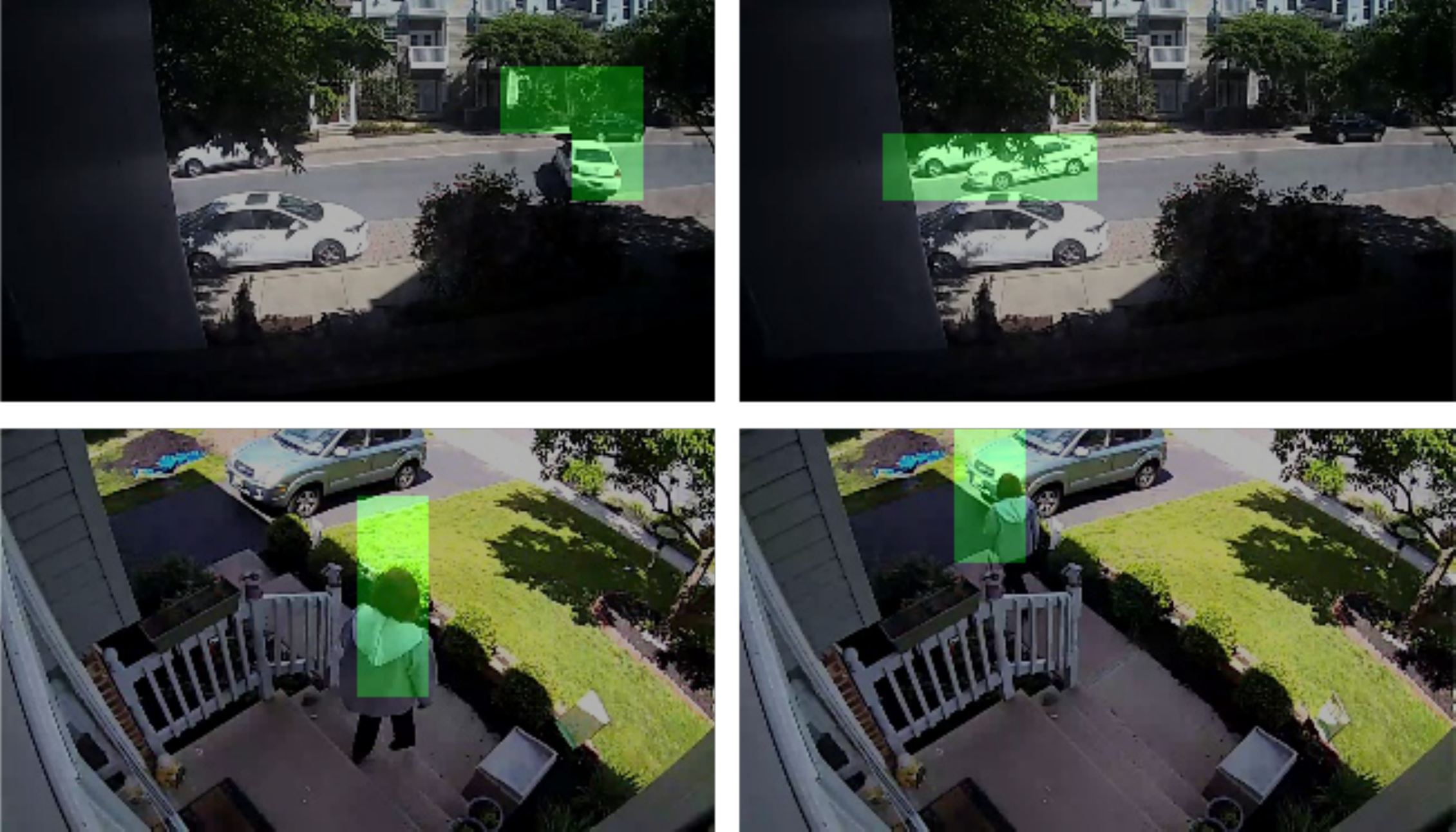}
  \caption{Pseudo-groundtruth Obtained from the Object Detection based Method.}
\label{fig:PseuGT}
\end{figure}

To map the mask with original resolution (1280$\times$720) to the pseudo-groundtruth of the STA layer, which has a much smaller spatial size due to spatial-wise max-pooling layers, we divide the original frame into 5$\times$3 grids (if choosing 160$\times$90 as the input resolution), which is the spatial size of the output responses from the STA layer in our network. The size of each grid is 256$\times$240.
For each grid, if more than half of its pixels are labeled as ``1" in the binary mask of the original resolution, we label this grid as ``1". Finally, we obtain a 5$\times$3 binary mask for each frame as the pseudo-groundtruth to train the STA layer. We show some samples of the obtained pseudo-groundtruth in Figure \ref{fig:PseuGT}.

{\small
\bibliographystyle{ieee}
\bibliography{egbib}
}

\end{document}